\documentclass[10pt,twocolumn,letterpaper]{article}

\usepackage{cvpr}              

\usepackage{graphicx}
\usepackage{amsmath}
\usepackage{amssymb}
\usepackage{booktabs}

\usepackage[dvipsnames]{xcolor}
\usepackage{subcaption}
\usepackage{booktabs}
\usepackage{multirow}
\usepackage{colortbl} 
\usepackage{pifont}  
\usepackage{cite}  
\usepackage{mathtools}  
\usepackage[font=small,labelfont=bf]{caption}  
\usepackage{blindtext}  
\usepackage{bbm}  
\usepackage[accsupp]{axessibility}
\usepackage[pagebackref,breaklinks,colorlinks]{hyperref}

\newcommand{\ournet}{ASNet\xspace}
\newcommand{\ourmethod}{iFSL\xspace}
\newcommand{\fsctask}{FS-C\xspace}
\newcommand{\fsstask}{FS-S\xspace}
\newcommand{\ourtask}{FS-CS\xspace}
\newcommand{\ourlayer}{AS\xspace}

\newcommand{\textcls}{\text{C}}
\newcommand{\textseg}{\text{S}}
\newcommand{\textgt}{\text{gt}}
\newcommand{\Real}{\mathbb{R}}

\newcommand{\bC}{\mathbf{C}}

\newcommand{\bp}{\mathbf{p}}
\newcommand{\bF}{\mathbf{F}}
\newcommand{\bA}{\mathbf{A}}

\newcommand{\bx}{\mathbf{x}}

\newcommand{\bV}{\mathbf{V}}

\newcommand{\bW}{\mathbf{W}}

\newcommand{\bY}{\mathbf{Y}}
\newcommand{\by}{\mathbf{y}}
\newcommand{\bK}{\mathbf{K}}

\newcommand{\bT}{\mathbf{T}}
\newcommand{\texts}{{\text{s}}}
\newcommand{\textq}{{\text{q}}}
\newcommand{\textk}{{\text{k}}}

\newcommand{\texto}{{\text{o}}}
\newcommand{\textA}{{\text{A}}}
\newcommand{\textI}{{\text{I}}}
\newcommand{\textt}{{\text{t}}}

\newcommand{\gray}[1]{\textcolor{gray}{#1}}

\newcommand{\conditionalcomment}[1]{\if\commenttext1 \else {#1} \fi}
\newcommand{\grayconditionalcomment}[1]{\if\commenttext1 \else \gray{{#1}} \fi}

\newcommand{\smallbreakparagraph}[1]{\smallbreak \noindent \textbf{#1}}

\newcommand{\figureref}[2]{Figure~\ref{#1}\xspace}
\newcommand{\figref}[1]{Fig.~\ref{#1}\xspace}
\newcommand{\tableref}[1]{Table~\ref{#1}}

\renewcommand{\eqref}[1]{Eq.~(\ref{#1})}

\DeclareMathOperator*{\argmax}{arg\,max}

\definecolor{grey}{rgb}{0.9, 0.9, 0.9}
\newcommand{\ccol}{\cellcolor{grey}}

\usepackage[capitalize]{cleveref}
\crefname{section}{Sec.}{Secs.}
\Crefname{section}{Section}{Sections}
\Crefname{table}{Table}{Tables}
\crefname{table}{Tab.}{Tabs.}

\def\cvprPaperID{750} 
\def\confName{CVPR}
\def\confYear{2022}

\begin{document}

\title{Integrative Few-Shot Learning for Classification and Segmentation}

\author{Dahyun Kang \quad Minsu Cho\vspace{0.15cm}\\
Pohang University of Science and Technology (POSTECH), South Korea\\
{
\small
\small \url{http://cvlab.postech.ac.kr/research/iFSL}
}
}
\maketitle


\begin{abstract}
We introduce the integrative task of few-shot classification and segmentation (\ourtask) 
that aims to both classify and segment target objects in a query image when the target classes are given with a few examples.  
This task combines two conventional few-shot learning problems, few-shot classification and segmentation.
\ourtask generalizes them to more realistic episodes with arbitrary image pairs, where each target class may or may not be present in the query.
To address the task, we propose the integrative few-shot learning (\ourmethod) framework for \ourtask, 
which trains a learner to construct class-wise foreground maps for multi-label classification and pixel-wise segmentation.
We also develop an effective \ourmethod model, attentive squeeze network (\ournet), that leverages deep semantic correlation and global self-attention to produce reliable foreground maps.
In experiments, the proposed method shows promising performance on the \ourtask task and also achieves the state of the art on standard few-shot segmentation benchmarks.
\end{abstract}


\begin{figure}[t!]
	\centering
	\small
    \includegraphics[width=\linewidth]{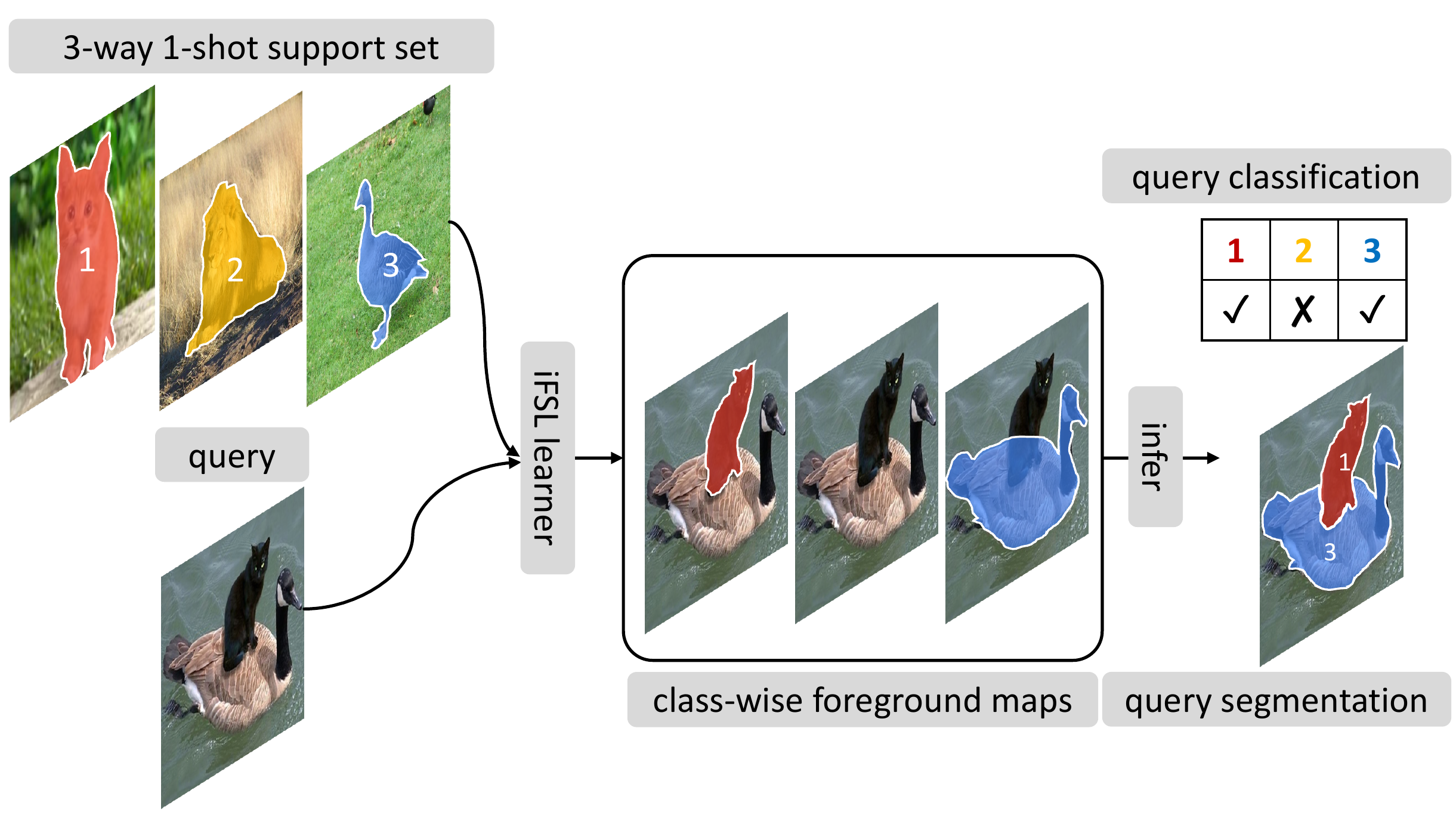}
    \hfill
    \includegraphics[width=\linewidth]{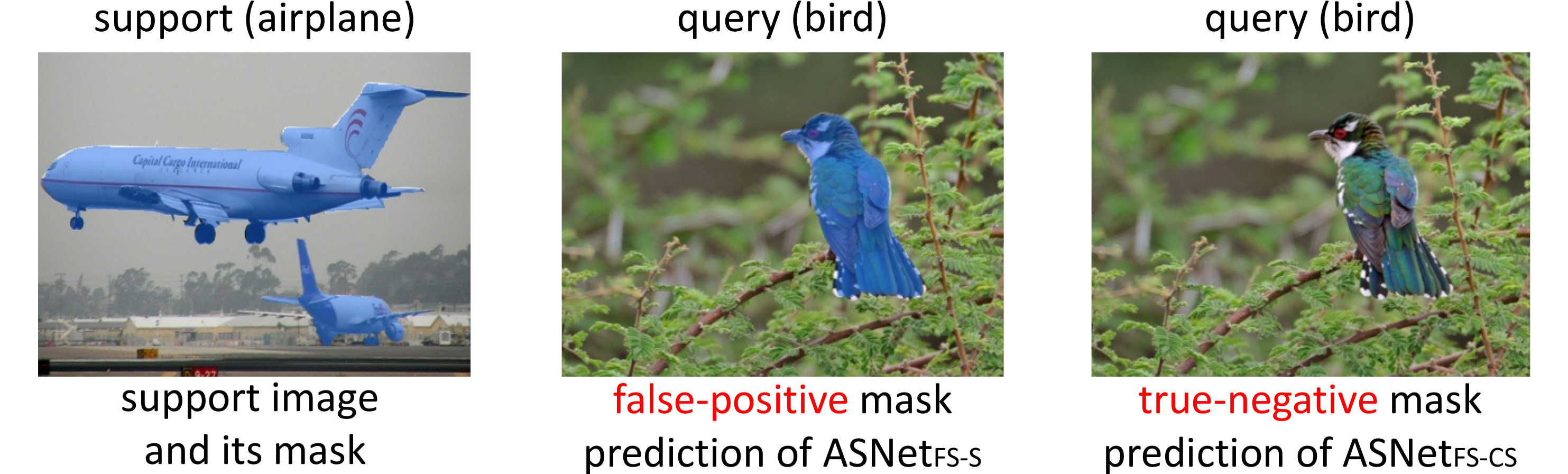}
	\caption{
	\textbf{Top:} Integrative few-shot learning framework (\ourmethod) for integrative few-shot classification and segmentation (\ourtask). 
    \textbf{Bottom:} 
    \fsstask learners are trained to segment a query image using a semantically-coupled support set thus often blindly highlight any salient objects regardless of support semantics.
    The proposed \ourtask learners are trained to predict class presence as well as corresponding masks thus correctly discriminate what to segment based on the semantic relevance between the query and the support.
    }
    \vspace{-4mm}
\label{fig:teaser}
\end{figure}

\section{Introduction}
Few-shot learning~\cite{fei2006one, fink2005object, wu2010towards, lake2015human, wang2020generalizing} is the learning problem where a learner experiences only a limited number of examples as supervision.
In computer vision, it has been most actively studied for the tasks of image classification~\cite{alexnet, vgg, resnet} and semantic segmentation~\cite{deeplab, fcn, deconvnet, unet} among many others~\cite{han2021query, ojha2021few, ramon2021h3d, yue2021prototypical, zhao2021few}.
Few-shot classification (\fsctask) aims to classify a query image into target classes when a few support examples are given for each target class. 
Few-shot segmentation (\fsstask) is to segment out the target class regions on the query image in a similar setup.
While being closely related to each other~\cite{li2009towards, yao2012describing, zhou2019collaborative},
these two few-shot learning problems have so far been treated individually.
Furthermore, the conventional setups for the few-shot problems, \fsctask and \fsstask, are limited and do not reflect realistic scenarios; 
\fsctask~\cite{matchingnet, ravi2016optimization, koch2015siamese} presumes that the query always contains one of the target classes in classification, while \fsstask~\cite{shaban2017oslsm, rakelly2018cofcn, hu2019amcg} allows the presence of multiple classes but does not handle the absence of the target classes in segmentation.
These respective limitations prevent few-shot learning from generalizing to and evaluating on more realistic cases in the wild. 
For example, when a query image without any target class is given as in \figref{fig:teaser}, \fsstask learners typically segment out arbitrary salient objects in the query.

To address the aforementioned issues, we introduce the \textit{integrative task of few-shot classification and segmentation} (\ourtask) that combines the two few-shot learning problems into a multi-label and background-aware prediction problem. 
Given a query image and a few-shot support set for target classes,  \ourtask aims to \textit{identify the presence of each target class} and \textit{predict its foreground mask} from the query.
Unlike \fsctask and \fsstask, it does not presume either the class exclusiveness in classification or the presence of all the target classes in segmentation.

As a learning framework for \ourtask, we propose {\em integrative few-shot learning} (\ourmethod) that learns to construct shared foreground maps for both classification and segmentation.
It naturally combines multi-label classification and pixel-wise segmentation by sharing class-wise foreground maps and also allows to learn with class tags or segmentation annotations. 
For effective \ourmethod, we design the {\em attentive squeeze network} (\ournet) that computes semantic correlation tensors between the query and the support image features and then transforms the tensor into a foreground map by strided self-attention. 
It generates reliable foreground maps for \ourmethod by leveraging multi-layer neural features~\cite{hpf, hsnet} and global self-attention~\cite{transformers, vit}. 
In experiments, we demonstrate the efficacy of the \ourmethod framework on \ourtask and compare \ournet with recent methods~\cite{xie2021few, wu2021learning, hsnet, xie2021scale}.
Our method significantly improves over the other methods on \ourtask in terms of classification and segmentation accuracy and also outperforms the recent \fsstask methods on the conventional \fsstask.
We also cross-validate the task transferability between the \fsctask, \fsstask, and \ourtask learners, and show the \ourtask learners effectively generalize when transferred to the \fsctask and \fsstask tasks.

Our contribution is summarized as follows:
\begin{itemize}
    \item We introduce the task of \textit{integrative few-shot classification and segmentation} (\ourtask), which combines few-shot classification and few-shot segmentation into an integrative task by addressing their limitations.
    \item We propose the \textit{integrative few-shot learning framework} (\ourmethod), which learns to both classify and segment a query image using class-wise foreground maps.
    \item We design the \textit{attentive squeeze network} (\ournet),
    which squeezes semantic correlations into a foreground map for \ourmethod via strided global self-attention.
    \item We show in extensive experiments that the framework, \ourmethod, and the architecture, \ournet, are both effective, achieving a significant gain on \fsstask as well as \ourtask.
\end{itemize}

\section{Related work}
\smallbreakparagraph{Few-shot classification (\fsctask)}.
Recent \fsctask methods typically learn neural networks that maximize positive class similarity and suppress the rest to predict the most probable class.
Such a similarity function is obtained by a) meta-learning embedding functions~\cite{koch2015siamese, matchingnet, protonet, allen2019infinite, tewam, can, feat, deepemd, renet}, b) meta-learning to optimize classifier weights~\cite{maml, leo, mtl}, or c) transfer learning~\cite{closer, rfs, dhillon2019baseline, wang2020few, negmargin, gidaris2018dynamic, qi2018low, rodriguez2020embedding}, all of which aim to generalize to unseen classes.
This conventional formulation is applicable if a query image corresponds to no less or more than a single class among target classes.
To generalize \fsctask to classify images associated with either none or multiple classes,
we employ the multi-label classification~\cite{mccallum1999multi, boutell2004learning, cole2021multi, lanchantin2021general, durand2019learning}.
While the conventional \fsctask methods make use of the class uniqueness property via using the categorical cross-entropy, we instead devise a learning framework that compares the binary relationship between the query and each support image individually and estimates a binary presence of the corresponding class.

\smallbreakparagraph{Few-shot semantic segmentation (\fsstask)}.
A prevalent \fsstask approach is learning to match a query feature map with a set of support feature embeddings that are obtained by collapsing spatial dimensions at the cost of spatial structures~\cite{wang2019panet, zhang2021self, siam2019amp, yang2021mining, liu2021anti, dong2018few, nguyen2019fwb, zhang2019canet, gairola2020simpropnet, yang2020pmm, liu2020ppnet}.
Recent methods~\cite{zhang2019pgnet, xie2021scale, xie2021few, wu2021learning, tian2020pfenet} focus on learning structural details by leveraging dense feature correlation tensors between the query and each support.
HSNet~\cite{hsnet} learns to squeeze a dense feature correlation tensor and transform it to a segmentation mask via high-dimensional convolutions that analyze the local correlation patterns on the correlation pyramid.
We inherit the idea of learning to squeeze correlations and improve it by analyzing the spatial context of the correlation with effective global self-attention~\cite{transformers}.  
Note that several methods~\cite{yang2020brinet, wang2020dan, sun2021boosting} adopt non-local self-attention~\cite{nlsa} of the query-key-value interaction for \fsstask, but they are distinct from ours in the sense that they learn to transform image feature maps, whereas our method focuses on transforming dense correlation maps via self-attention.

\fsstask has been predominantly investigated as an one-way segmentation task, \ie, foreground or background segmentation, since the task is defined so that every target (support) class object appears in query images, thus being not straightforward to extend to a multi-class problem in the wild.
Consequently, most work on \fsstask except for a few~\cite{wang2019panet, tian2020differentiable, liu2020ppnet, dong2018few} focuses on the one-way segmentation, where the work of \cite{tian2020differentiable, dong2018few} among the few presents two-way segmentation results from person-and-object images only, \eg, images containing (person, dog) or (person, table).

\smallbreakparagraph{Comparison with other few-shot approaches.}
Here we contrast \ourtask with other loosely-related work for generalized few-shot learning.
Few-shot open-set classification~\cite{liu2020few} brings the idea of the open-set problem~\cite{scheirer2012toward, fei2016breaking} to few-shot classification by allowing a query to have no target classes.
This formulation enables background-aware classification as in \ourtask, whereas multi-label classification is not considered.
The work of \cite{tian2020generalized, ganea2021incremental} generalizes few-shot segmentation to a multi-class task, but it is mainly studied under the umbrella of incremental learning~\cite{mccloskey1989catastrophic, rebuffi2017icarl, castro2018end}.
The work of \cite{siam2020weakly} investigates weakly-supervised few-shot segmentation using image-level vision and language supervision, while \ourtask uses visual supervision only.
The aforementioned tasks generalize few-shot learning but differ from \ourtask in the sense that \ourtask integrates two related problems under more general and relaxed constraints.

\section{Problem formulation}
\label{sec:ourtask}
Given a query image and a few support images for target classes, we aim to {\em identify the presence} of each class and {\em predict its foreground mask} from the query (\figref{fig:teaser}), which we call the {\em integrative few-shot classification and segmentation} (\ourtask).  
Specifically, let us assume a target (support) class set $\mathcal{C}_\texts$ of $N$ classes and its support set $\mathcal{S}=\{ (\bx_\texts^{(i)}, y_\texts^{(i)}) | y_\texts^{(i)} \in \mathcal{C}_\texts \}^{NK}_{i=1}$, which contains $K$ labeled instances for each of the $N$ classes, \ie, $N$-way $K$-shot~\cite{matchingnet, ravi2016optimization}. 
The label $y_\texts^{(i)}$ is either a class tag (weak label) or a segmentation annotation (strong label). 
For a given query image $\bx$, we aim to identify the multi-hot class occurrence $\by_\textcls$ and also predict the segmentation mask $\bY_\textseg$ corresponding to the classes. 
We assume the class set of the query $\mathcal{C}$ is a subset of the target class set, \ie, $\mathcal{C} \subseteq \mathcal{C}_\texts$, thus it is also possible to obtain $\by_\textcls = \varnothing$ and $\bY_\textseg = \varnothing$. 
This naturally generalizes the existing few-shot classification~\cite{matchingnet, protonet} and few-shot segmentation~\cite{shaban2017oslsm, rakelly2018cofcn}.

\smallbreakparagraph{Multi-label background-aware prediction.}
The conventional formulation of few-shot classification (\fsctask)~\cite{matchingnet, protonet, maml} assigns the query to one class among the target classes exclusively and ignores the possibility of the query belonging to none or multiple target classes. 
\ourtask tackle this limitation and generalizes \fsctask to multi-label classification with a background class. 
A multi-label few-shot classification learner $f_{\textcls}$ compares semantic similarities between the query and the support images and estimates class-wise occurrences: $\hat{\by}_{\textcls} = f_{\textcls}(\bx, \mathcal{S}; \theta)$ where $\hat{\by}_{\textcls}$ is an $N$-dimensional multi-hot vector each entry of which indicates the occurrence of the corresponding target class.
Note that the query is classified into a \textit{background} class if none of the target classes were detected.  
Thanks to the relaxed constraint on the query, \ie, the query not always belonging to exactly one class, \ourtask is more general than \fsctask.

\smallbreakparagraph{Integration of classification and segmentation.}
\ourtask integrates multi-label few-shot classification with semantic segmentation by adopting pixel-level spatial reasoning.
While the conventional \fsstask~\cite{shaban2017oslsm, rakelly2018cofcn, wang2019panet, siam2019amp, nguyen2019fwb} assumes the query class set exactly matches the support class set,
\ie, $\mathcal{C} = \mathcal{C}_\texts$,
\ourtask relaxes the assumption such that the query class set can be a subset of the support class set,
\ie, $\mathcal{C} \subseteq \mathcal{C}_\texts$.
In this generalized segmentation setup along with classification, an integrative \ourtask learner $f$ estimates both class-wise occurrences and their semantic segmentation maps: $\{ \hat{\by}_{\textcls}, \hat{\bY}_{\textseg}\} = f(\bx, \mathcal{S} ; \theta)$.
This combined and generalized formulation gives a high degree of freedom to both of the few-shot learning tasks, which has been missing in the literature;
the integrative few-shot learner can predict multi-label background-aware class occurrences and segmentation maps simultaneously under a relaxed constraint on the few-shot episodes.

\begin{figure*}[t!]
	\centering
	\small
    \includegraphics[width=\linewidth]{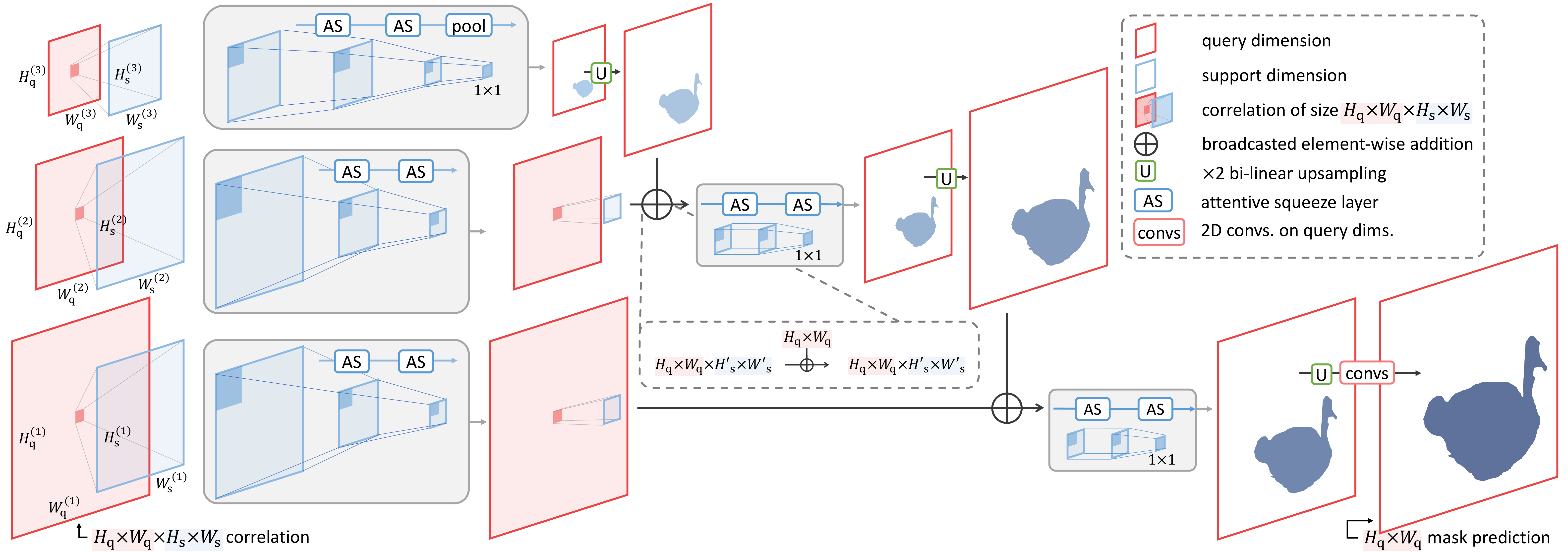}
	\caption{\textbf{Overview of \ournet.} 
    \ournet first constructs a hypercorrelation~\cite{hsnet} with image feature maps between a query (colored red) and a support (colored blue), where the 4D correlation is depicted as two 2D squares for demonstrational simplicity.
    \ournet then learns to transform the correlation to a foreground map by gradually squeezing the support dimension on each query dimension via global self-attention.
    Each input correlation, intermediate feature, and output foreground map has a channel dimension but is omitted in the illustration.
}
\label{fig:overview}
\end{figure*}

\section{Integrative Few-Shot Learning (\ourmethod)}
\label{sec:ourmethod}
To solve the \ourtask problem, we propose an effective learning framework, \textit{integrative few-shot learning (\ourmethod)}.
The \ourmethod framework is designed to jointly solve few-shot classification and few-shot segmentation using either a class tag or a segmentation supervision.
The integrative few-shot learner $f$ takes as input the query image $\bx$ and the support set $\mathcal{S}$ and then produces as output the class-wise foreground maps.
The set of class-wise foreground maps $\mathcal{Y}$ is comprised of $\bY^{(n)} \in \Real^{H \times W}$ for $N$ classes: 
\begin{align}
\mathcal{Y} = f(\bx, \mathcal{S}; \theta) = \{ \bY^{(n)}\}_{n=1}^{N},
\label{eq:foreground_mask}
\end{align}
where $H \times W$ denotes the size of each map and $\theta$ is parameters to be meta-learned.
The output at each position on the map represents the probability of the position being on a foreground region of the corresponding class.

\smallbreakparagraph{Inference.}
\ourmethod infers both class-wise occurrences and segmentation masks on top of the set of foreground maps $\mathcal{Y}$.
For class-wise occurrences, a multi-hot vector $\hat{\by}_\textcls \in \Real^{N}$ is predicted via max pooling followed by thresholding:
\begin{align}
\hat{\by}_{\textcls}^{(n)} &= 
\begin{cases}
  1  \text{\; if \,} \max_{\bp \in [H] \times [W]} \bY^{(n)}(\bp) \geq \delta,\\
  0 \text{\; otherwise,}
\end{cases}
\label{eq:predict_class}
\end{align}
where $\bp$ denotes a 2D position, $\delta$ is a threshold, and $[ k ]$ denotes a set of integers from 1 to $k$, \ie, $[k] = \{1,\! 2,\! \cdots,\! k\}$.
We find that inference with average pooling is prone to miss small objects in multi-label classification and thus choose to use max pooling.
The detected class at any position on the spatial map signifies the presence of the class.  

For segmentation, a segmentation probability tensor $\bY_{\textseg} \in \Real^{ H \times W \times (N + 1)}$ is derived from the class-wise foreground maps.
As the background class is not given as a separate support, we estimate the background map in the context of the given supports;  we combine $N$ class-wise background maps into \textit{an episodic background map} on the fly. 
Specifically, we compute the episodic background map $\bY_{\text{bg}}$ by averaging the probability maps of not being foreground and then concatenate it with the class-wise foreground maps to obtain a segmentation probability tensor $\bY_{\textseg}$: 
\begin{align}
\bY_{\text{bg}} &= \frac{1}{N} \sum_{n=1}^{N}(\mathbf{1} - \bY^{(n)}), \label{eq:bg_mask}\\
\bY_{\textseg} &= \left[ \bY || \bY_{\text{bg}} \right] \in \Real^{ H \times W \times (N + 1)}.
\label{eq:merge_mask}
\end{align}
The final segmentation mask $\hat{\bY}_\textseg \in \Real^{H \times W}$ is obtained by computing the most probable class label for each position:
\begin{equation}
\hat{\bY}_\textseg = \argmax_{n \in [N + 1]} \bY_{\textseg}.
\label{eq:predict_mask}
\end{equation}

\smallbreakparagraph{Learning objective.}
The \ourmethod framework allows a learner to be trained using a class tag or a segmentation annotation using the classification loss or segmentation loss, respectively.
The classification loss is formulated as the average binary cross-entropy between the spatially average-pooled class scores and its ground-truth class label:
\begin{align}
    \mathcal{L}_{\textcls} &= -\frac{1}{N}\sum_{n=1}^{N}\by_{\textgt}^{(n)} \log \frac{1}{H W}\sum_{\scriptscriptstyle{\bp \in [H] \! \times \! [W]}} \bY^{(n)}(\bp), 
        \label{eq:loss_cls_final}
\end{align}
where $\by_\textgt$ denotes the multi-hot encoded ground-truth class.

The segmentation loss is formulated as the average cross-entropy between the class distribution at each individual position and its ground-truth segmentation annotation:
\begin{align}
    \mathcal{L}_{\textseg} &= - \frac{1}{(N + 1)}\frac{1}{H W} \sum_{n=1}^{N + 1}\sum_{\scriptscriptstyle{\bp \in [H] \! \times \! [W]}} \bY_\textgt^{(n)}(\bp) \log \bY_{\textseg}^{(n)}(\bp), 
        \label{eq:loss_seg_final}
\end{align} where $\bY_\textgt$ denotes the ground-truth segmentation mask. 

These two losses share a similar goal of classification but differ in whether to classify each \textit{image} or each \textit{pixel}. Either of them is thus chosen according to the given level of supervision for training.

\section{Model architecture}
In this section, we present \textit{Attentive Squeeze Network} (\ournet) of an effective \ourmethod model.
The main building block of \ournet is the attentive squeeze layer (\ourlayer layer), which is a high-order self-attention layer that takes a correlation tensor and returns another level of correlational representation.
\ournet takes as input the pyramidal cross-correlation tensors between a query and a support image feature pyramids, \ie, a hypercorrelation~\cite{hsnet}.
The pyramidal correlations are fed to pyramidal \ourlayer layers that gradually squeeze the spatial dimensions of the support image, and the pyramidal outputs are merged to a final foreground map in a bottom-up pathway~\cite{hsnet, fpn, refinenet}.
\figureref{fig:overview}~illustrates the overall process of \ournet.
The $N$-way output maps are computed in parallel and collected to prepare the class-wise foreground maps in \eqref{eq:foreground_mask} for \ourmethod.

\subsection{Attentive Squeeze Network (ASNet)}
\smallbreakparagraph{Hypercorrelation construction.}
Our method first constructs $NK$ hypercorrelations~\cite{hsnet} between a query and each $NK$ support image and then learns to generate a foreground segmentation mask \wrt each support input.
To prepare the input hypercorrelations, an episode, \ie, a query and a support set, is enumerated into a paired list of the query, a support image, and a support label: $\{(\bx, (\bx_\texts^{(i)}, y_\texts^{(i)})) \}_{i=1}^{NK}$.
The input image is fed to stacked convolutional layers in a CNN and its mid- to high-level output feature maps are collected to build a feature pyramid $\{\bF^{(l)}\}_{l=1}^{L}$, where $l$ denotes the index of a unit layer, \eg, $\texttt{Bottleneck}$ layer in ResNet50~\cite{resnet}.
We then compute cosine similarity between each pair of feature maps from the pair of query and support feature pyramids to obtain 4D correlation tensors of size $H_\textq^{(l)} \times W_\textq^{(l)} \times H_\texts^{(l)} \times W_\texts^{(l)}$, which is followed by ReLU~\cite{relu}:
\begin{equation}
\bC^{(l)}(\bp_\textq, \bp_\texts) = \mathrm{ReLU}\left( \frac{\bF_\textq^{(l)}(\bp_\textq) \cdot \bF_\texts^{(l)}(\bp_\texts)}{||\bF_\textq^{(l)}(\bp_\textq)|| \, ||\bF_\texts^{(l)}(\bp_\texts)||} \right).
\end{equation}
These $L$ correlation tensors are grouped by $P$ groups of the identical spatial sizes, and then the tensors in each group are concatenated along a new channel dimension to build a hypercorrelation pyramid: $\{\bC^{(p)} | \bC^{(p)} \in \Real^{ H_\textq^{(p)} \times W_\textq^{(p)} \times H_\texts^{(p)} \times W_\texts^{(p)} \times C_{\text{in}}^{(p)}}  \}_{p=1}^{P}$ such that the channel size $C_{\text{in}}^{(p)}$ corresponds to the number of concatenated tensors in the $p_{\text{th}}$ group. 
We denote the first two spatial dimensions of the correlation tensor, \ie, $\Real^{H_\textq \times W_\textq}$, as query dimensions, and the last two spatial dimensions, \ie, $\Real^{H_\texts \times W_\texts}$, as support dimensions hereafter.

\smallbreakparagraph{Attentive squeeze layer (\ourlayer layer).}
The \ourlayer layer transforms a correlation tensor to another with a smaller support dimension via strided self-attention. 
The tensor is recast as a matrix with each element representing a support pattern.
Given a correlation tensor $\bC \in \Real^{H_\textq \times W_\textq \times H_\texts \times W_\texts \times C_{\text{in}} }$ in a hypercorrelation pyramid, we start by reshaping the correlation tensor as a block matrix of size $H_\textq \times W_\textq$ with each element corresponding to a correlation tensor of $\bC(\bx_\textq) \in \Real^{H_\texts \times W_\texts \times C_{\text{in}} }$ on the query position $\bx_\textq$ such that 
\begin{equation}
\bC^{\text{block}} = 
\begin{bmatrix} 
    \bC((1, 1)) & \hdots & \bC((1, W_\textq)) \\
    \vdots & \ddots & \vdots   \\
    \bC((H_\textq, 1)) &  \hdots & \bC((H_\textq, W_\textq)) 
\end{bmatrix}
\label{eq:block_matrix}
.
\end{equation}
We call each element a \textit{support correlation tensor}.
The goal of an \ourlayer layer is to analyze the global context of each support correlation tensor and extract a correlational representation with a reduced support dimension while the query dimension is preserved: 
$\Real^{ H_\textq \times W_\textq \times H_\texts \times W_\texts \times C_{\text{in}} } \rightarrow \Real^{H_\textq \times W_\textq \times H_\texts' \times W_\texts' \times C_{\text{out}} }$, where $H_\texts' \leq H_\texts$ and $W_\texts' \leq W_\texts$.
To learn a holistic pattern of each support correlation, we adopt the global self-attention mechanism~\cite{transformers} for correlational feature transform.
The self-attention weights are shared across all query positions and processed in parallel.

Let us denote a support correlation tensor on any query position $\bx_\textq$ by  $\bC^{\texts} = \bC^{\text{block}}(\bx_\textq)$ for notational brevity as all positions share the following computation.
The self-attention computation starts by embedding a support correlation tensor $\bC^{\texts}$ to a target
\footnote{
In this section, we adapt the term ``target'' to indicate the ``query'' embedding in the context of self-attention learning~\cite{transformers, vit, lsa, pvt, lrnet} to avoid homonymous confusion with the ``query'' image to be segmented.
}
, key, value triplet: 
$
\bT, \bK, \bV \in \Real^{ H_\texts' \times W_\texts' \times C_{\text{hd}}},
$
using three convolutions of which strides greater than or equal to one to govern the output size.
The resultant target and key correlational representations, $\bT$ and $\bK$, are then used to compute an attention context.
The attention context is computed as following matrix multiplication:
\begin{equation}
\bA = \bT \bK^{\top} \in \Real^{H_\texts' \times W_\texts' \times H_\texts' \times W_\texts'}.
\label{eq:sixd_attn}
\end{equation}
Next, the attention context is normalized by softmax such that the votes on key foreground positions sum to one with masking attention by the support mask annotation $\bY_\texts$ if available to attend more on the foreground region:
\begin{align*}
\bar{\bA}(\bp_\textt, \bp_\textk) = \frac{\exp \left( \bA(\bp_\textt, \bp_\textk) \bY_\texts(\bp_\textk) \right)}{\sum_{\bp_\textk'}\exp \left( \bA(\bp_\textt, \bp'_\textk) \bY_\texts(\bp'_\textk) \right)},
\end{align*}
\begin{equation}
\text{where\, } \bY_\texts(\bp_\textk) =
\begin{cases}
  1 \quad \; \; \text{if} \; \bp_\textk \in [H'_\texts] \times [W'_\texts] \text{ is foreground,} \\    
  - \infty  \; \text{otherwise.}
\end{cases} 
\label{eq:masked_attn}
\end{equation}
The masked attention context $\bar{\bA}$ is then used to aggregate
the value embedding $\bV$:
\begin{equation}
\bC^{\texts}_{\textA} = \bar{\bA} \bV \in \Real^{ H_\texts' \times W_\texts' \times C_{\text{hd}} }.
\label{eq:agg}
\end{equation}
The attended representation is fed to an MLP layer, $\bW_\texto$, and added to the input.
In case the input and output dimensions mismatch, the input is optionally fed to a convolutional layer, $\bW_\textI$.
The addition is followed by an activation layer $\varphi(\cdot)$ consisting of a group normalization~\cite{groupnorm} and a ReLU activation~\cite{relu}:
\begin{equation}
\bC^{\texts}_\texto = \varphi(\bW_\texto(\bC^{\texts}_\textA) + \bW_\textI(\bC^{\texts}) ) \in \Real^{  H_\texts' \times W_\texts' \times C_{\text{out}}}.
\end{equation}
The output is then fed to another MLP that concludes a unit operation of an \ourlayer layer:
\begin{equation}
\bC^{\texts \prime} = \varphi(\bW_\text{FF}(\bC^{\texts}_\texto) + \bC^{\texts}_\texto) \in \Real^{ H_\texts' \times W_\texts' \times C_{\text{out}}},
\end{equation}
which is embedded to the corresponding query position in the block matrix of \eqref{eq:block_matrix}.
Note that the \ourlayer layer can be stacked to progressively reduce the size of support correlation tensor, $H_\texts' \times W_\texts'$, to a smaller size.
The overall pipeline of \ourlayer layer is illustrated in the supplementary material.

\smallbreakparagraph{Multi-layer fusion.}
The pyramid correlational representations are merged from the coarsest to the finest level by cascading a pair-wise operation of the following three steps: upsampling, addition, and non-linear transform.
We first bi-linearly upsample the bottommost correlational representation to the query spatial dimension of its adjacent earlier one and then add the two representations to obtain a mixed one $\bC^{\text{mix}}$.
The mixed representation is fed to two sequential \ourlayer layers until it becomes a point feature of size $H_\texts' = W_\texts'=1$, which is fed to the subsequent pyramidal fusion.
The output from the earliest fusion layer is fed to a convolutional decoder, which consists of interleaved 2D convolution and bi-linear upsampling that map the $C$-dimensional channel to 2 (foreground and background) and the output spatial size to the input query image size.
See \figref{fig:overview} for the overall process of multi-layer fusion.

\smallbreakparagraph{Class-wise foreground map computation.}
The $K$-shot output foreground activation maps are averaged to produce a mask prediction for each class.
The averaged output map is normalized by softmax over the two channels of the binary segmentation map to obtain a foreground probability prediction $\bY^{(n)} \in \Real^{H \times W}$.

\begin{table*}[t!]
    \centering
    \small
    \setlength{\tabcolsep}{5pt}
    \scalebox{0.80}{
        \begin{tabular}{lccccc|ccccc||ccccc|cccccr}
            \toprule
             & \multicolumn{10}{c}{1-way 1-shot} & \multicolumn{10}{c}{2-way 1-shot} \\
             \cmidrule(lr){2-11}\cmidrule(lr){12-21}
             & \multicolumn{5}{c}{classification 0/1 exact ratio (\%)} & \multicolumn{5}{c}{segmentation mIoU (\%)} & \multicolumn{5}{c}{classification 0/1 exact ratio (\%)} & \multicolumn{5}{c}{segmentation mIoU (\%)} \\
             \cmidrule(lr){2-6}\cmidrule(lr){7-11}\cmidrule(lr){12-16}\cmidrule(lr){17-21} method & $5^{0}$ & $5^{1}$ & $5^{2}$ & $5^{3}$ & avg. & $5^{0}$ & $5^{1}$ & $5^{2}$ & $5^{3}$ & avg. & $5^{0}$ & $5^{1}$ & $5^{2}$ & $5^{3}$ & avg. & $5^{0}$ & $5^{1}$ & $5^{2}$ & $5^{3}$ & avg. \\
             \midrule
             PANet \cite{wang2019panet} & 69.9 & 67.7 & 68.8 & 69.4 & 69.0 & 32.8 & 45.8 & 31.0 & 35.1 & 36.2 & 56.2 & 47.5 & 44.6 & 55.4 & 50.9 & 33.3 & 46.0 & 31.2 & 38.4 & 37.2 \\
             PFENet \cite{tian2020pfenet} & 69.8 & 82.4 & 68.1 & 77.9 & 74.6 & 38.3 & 54.7 & 35.1 & 43.8 & 43.0 & 22.5 & 61.7 & 40.3 & 39.5 & 41.0 & 31.1 & 47.3 & 30.8 & 32.2 & 35.3 \\
             HSNet \cite{hsnet} & \textbf{86.6} & 84.8 & 76.9 & \textbf{86.3} & 83.7 & 49.1 & 59.7 & 41.0 & 49.0 & 49.7 & 68.0 & 73.2 & 57.0 & \textbf{70.9} & 67.3 & 42.4 & 53.7 & 34.0 & 43.9 & 43.5 \\
             \ccol $\text{\ournet}_{\text{w}}$ & \ccol 86.4 & \ccol 86.3 & \ccol 70.9 & \ccol 84.5 & \ccol 82.0 & \ccol 10.8 & \ccol 20.2 & \ccol 13.1 & \ccol 16.1 & \ccol 15.0 & \ccol \textbf{71.6} & \ccol 72.4 & \ccol 46.4 & \ccol 68.0 & \ccol 64.6 & \ccol 11.4 & \ccol 20.8 & \ccol 12.5 & \ccol 15.9 & \ccol 15.1  \\
             \ccol \ournet & \ccol 84.9 & \ccol \textbf{89.6} & \ccol \textbf{79.0} & \ccol 86.2 & \ccol \textbf{84.9} & \ccol \textbf{51.7} & \ccol \textbf{61.5} & \ccol \textbf{43.3} & \ccol \textbf{52.8} & \ccol \textbf{52.3} & \ccol 68.5 & \ccol \textbf{76.2} & \ccol \textbf{58.6} & \ccol 70.0 & \ccol \textbf{68.3} & \ccol \textbf{48.5} & \ccol \textbf{58.3} & \ccol \textbf{36.3} & \ccol \textbf{48.3} & \ccol \textbf{47.8}  \\
             \bottomrule
        \end{tabular}
    }
    \vspace{-2mm}
    \caption{Performance comparison of \ournet and others on \ourtask and Pascal-5$^i$~\cite{shaban2017oslsm}.
    All methods are trained and evaluated under the \ourmethod framework given strong labels, \ie, class segmentation masks, except for $\text{\ournet}_{\text{w}}$ that is trained only with weak labels, \ie, class tags.
}
    \label{table:ipa}
    \vspace{-2mm}
\end{table*}

\begin{table}[t!]
    \centering
    \small
    \scalebox{1}{
        \begin{tabular}{lcc||cc}
            \toprule
             & \multicolumn{2}{c}{1-way 1-shot} & \multicolumn{2}{c}{2-way 1-shot}\\
             \cmidrule(lr){2-3}\cmidrule(lr){4-5}
             method & ER & mIoU & ER & mIoU \\
             \midrule
             PANet \cite{wang2019panet}   & 66.7 & 25.2 & 48.5 & 23.6 \\ 
             PFENet \cite{tian2020pfenet} & 71.4 & 31.9 & 36.5 & 22.6 \\ 
             HSNet \cite{hsnet}           & 77.0 & 34.3 & 62.5 & 29.5 \\ 
             \ccol \ournet             & \ccol \textbf{78.6} & \ccol \textbf{35.8} & \ccol \textbf{63.1} & \ccol \textbf{31.6} \\
             \bottomrule
        \end{tabular}
    }
    \vspace{-0mm}
    \caption{
    Performance comparison of \ournet and others on \ourtask and COCO-20$^i$~\cite{nguyen2019fwb}.
    }
    \label{table:ico}
    \vspace{-4mm}
\end{table}

\section{Experiments}
In this section we report our experimental results regarding the \ourtask task, the \ourmethod framework, as well as the  \ournet after briefly describing implementation details and evaluation benchmarks.
See the supplementary material for additional results, analyses, and experimental details.

\subsection{Experimental setups}
\smallbreakparagraph{Experimental settings.}
We select ResNet50 and ResNet-101~\cite{resnet} pretrained on ImageNet~\cite{russakovsky2015imagenet} as our backbone networks for a fair comparison with other methods and freeze the backbone during training as similarly as the previous work~\cite{tian2020pfenet, hsnet}.
We train models using Adam~\cite{adam} optimizer with learning rate of $10^{-4}$ and $10^{-3}$ for the classification loss and the segmentation loss, respectively.
We train all models with 1-way 1-shot training episodes and evaluate the models on arbitrary $N$-way $K$-shot episodes.
For inferring class occurrences, we use a threshold $\delta = 0.5$.
All the \ourlayer layers are implemented as multi-head attention with 8 heads. 
The number of correlation pyramid is set to $P=3$.

\smallbreakparagraph{Dataset.}
For the new task of \ourtask,
we construct a benchmark adopting the images and splits from the two widely-used \fsstask datasets, Pascal-5$^{i}$~\cite{shaban2017oslsm, pascal} and COCO-20$^{i}$~\cite{nguyen2019fwb, coco}, which are also suitable for multi-label classification~\cite{wang2017multi}.
Within each fold, we construct an episode by randomly sampling a query and an $N$-way $K$-shot support set that annotates the query with $N$-way class labels and an $(N\!+\!1)$-way segmentation mask in the context of the support set.
For the \fsstask task, we also use Pascal-5$^{i}$ and COCO-20$^{i}$ following the same data splits as \cite{shaban2017oslsm} and \cite{nguyen2019fwb}, respectively.

\smallbreakparagraph{Evaluation.}
Each dataset is split into four mutually disjoint class sets and cross-validated.
For multi-label classification evaluation metrics, we use the 0/1 exact ratio $\mathrm{ER} = \mathbbm{1} [\by_{\text{gt}} = \by_{\textcls} ]$~\cite{durand2019learning}.
In the supplementary material, we also report the results in accuracy $\mathrm{acc} = \frac{1}{N} \sum_n \mathbbm{1} [\by_{\text{gt}}^{(n)} =  \by^{(n)}_{\textcls} ]$.
For segmentation, we use mean IoU $\mathrm{mIoU} = \frac{1}{C}\sum_c \mathrm{IoU}_c$~\cite{shaban2017oslsm, wang2019panet}, where $\mathrm{IoU}_c$ denotes an IoU value of $c_{\text{th}}$ class.

\begin{figure}[t!]
	\centering
	\small
    \includegraphics[width=0.97\linewidth]{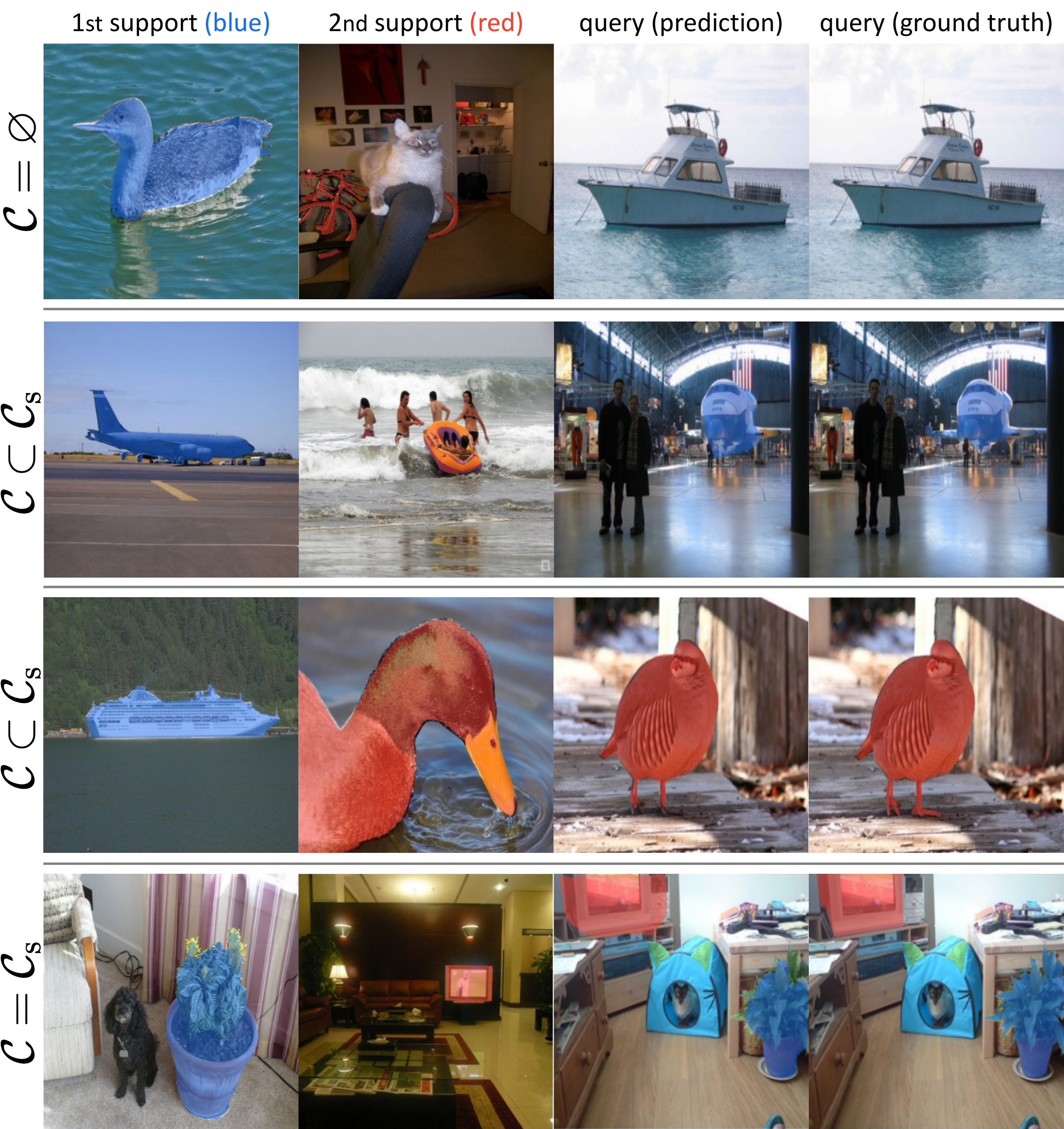}
    \vspace{-1mm}
	\caption{2-way 1-shot segmentation results of \ournet on \ourtask.
	The examples cover all three cases of $\mathcal{C}  =  \varnothing$, $\mathcal{C} \subset \mathcal{C}_\texts$, and $\mathcal{C} = \mathcal{C}_\texts$.
	The images are resized to square shape for visualization.
    \vspace{-4mm}
}
\label{fig:qual2way}
\end{figure}

\subsection{Experimental evaluation of \ourmethod on \ourtask}
In this subsection, we investigate the \ourmethod learning framework on the \ourtask task.
All ablation studies are conducted using ResNet50 on Pascal-$i^5$ and evaluated in 1-way 1-shot setup unless specified otherwise.
Note that it is difficult to present a fair and direct comparison between the conventional \fsctask and our few-shot classification task since \fsctask is always evaluated on single-label classification benchmarks~\cite{matchingnet, tieredimagenet, cifarfs, metaoptnet, metadataset}, whereas our task is evaluated on multi-label benchmarks~\cite{pascal, coco}, which are irreducible to a single-label one in general.

\smallbreakparagraph{Effectiveness of \ourmethod on \ourtask.}
We validate the \ourmethod framework on \ourtask and also compare the performance of \ournet with those of three recent state-of-the-art methods, PANet~\cite{wang2019panet}, PFENet~\cite{tian2020pfenet}, and HSNet~\cite{hsnet}, which are originally proposed for the conventional \fsstask task; all the models are trained by \ourmethod for a fair comparison.  
Note that we exclude the background merging step (Eqs.~\ref{eq:bg_mask} and \ref{eq:merge_mask}) for PANet as its own pipeline produces a multi-class output including background.
Tables \ref{table:ipa} and \ref{table:ico} validate the \ourmethod framework on the \ourtask task quantitatively, where our \ournet surpasses other methods on both 1-way and 2-way setups in terms of few-shot classification as well as the segmentation performance.
The 2-way segmentation results are also qualitatively demonstrated in \figref{fig:qual2way} visualizing exhaustive inclusion relations between a query class set $\mathcal{C}$ and a target (support) class set $\mathcal{C}_\texts$ in a 2-way setup.

\begin{figure}[t!]
	\centering
	\small
    \includegraphics[width=0.9\linewidth]{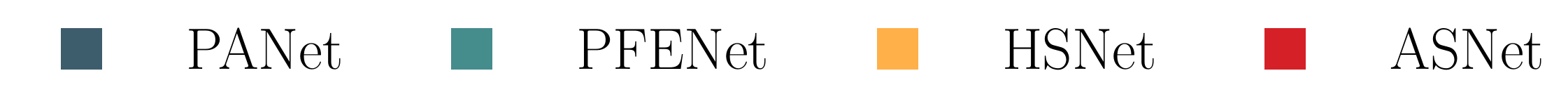}
    \vspace{-3mm}
    \includegraphics[width=0.49\linewidth]{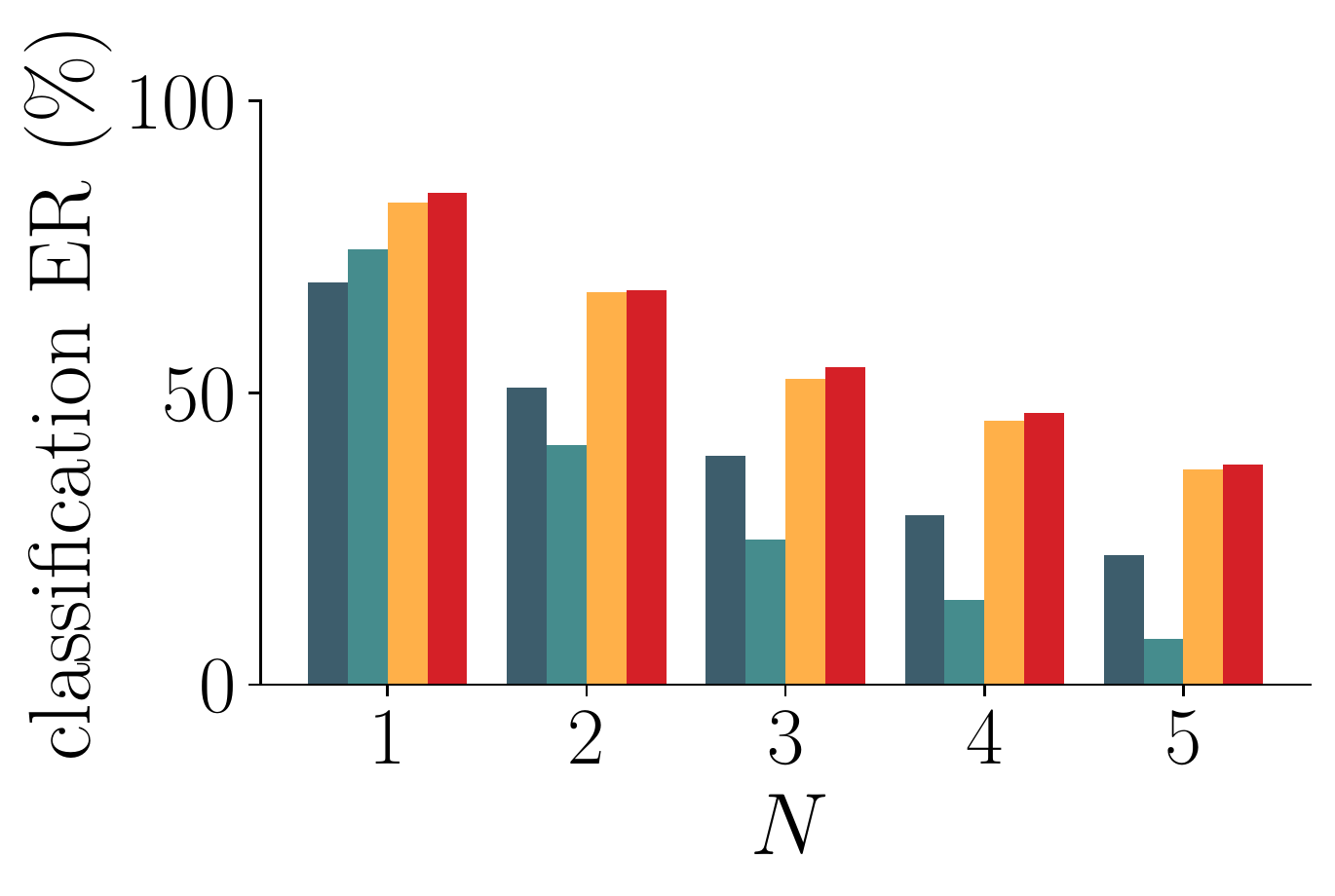} \hfill
    \includegraphics[width=0.49\linewidth]{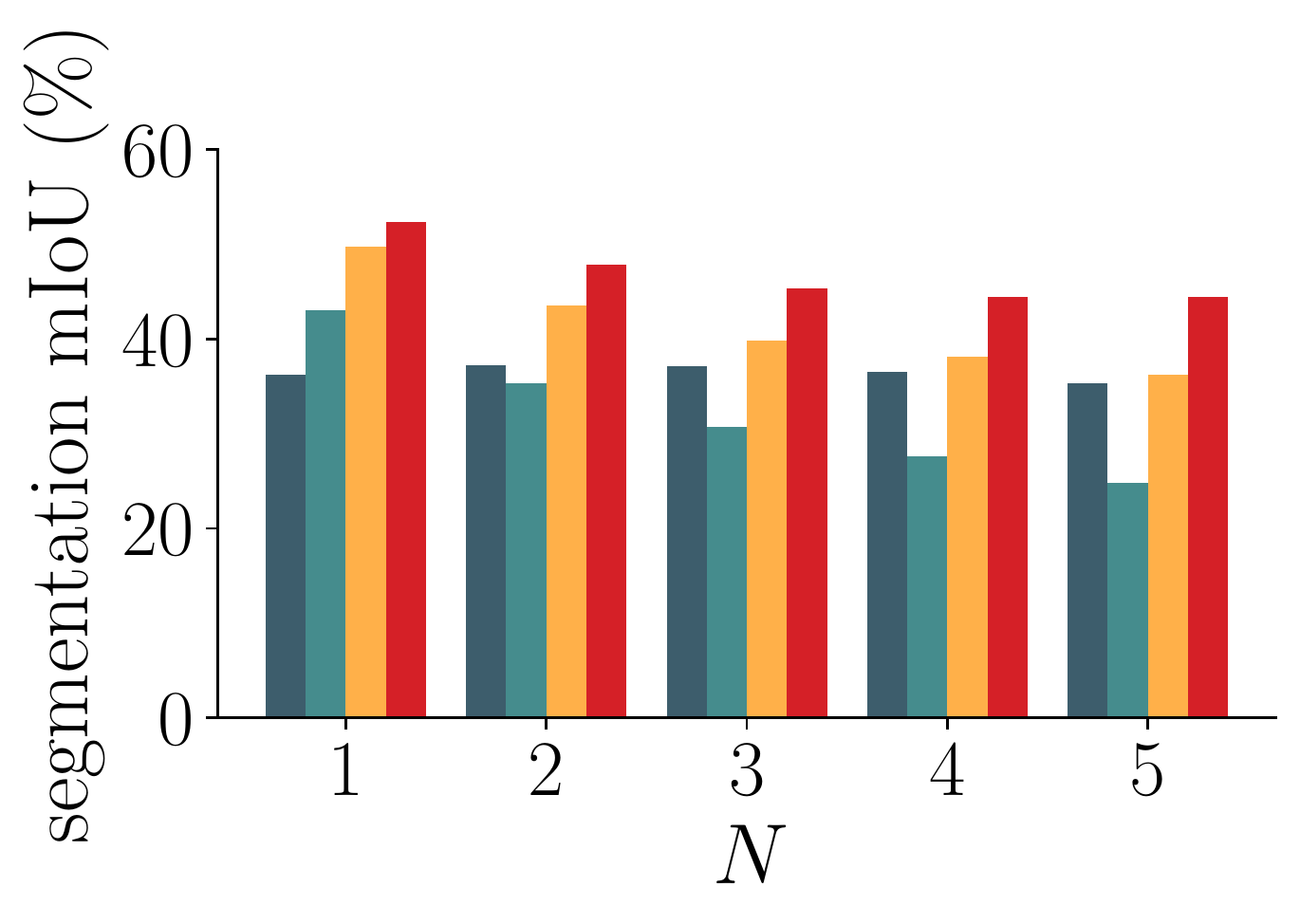}
    \caption{$N$-way 1-shot \ourtask performance comparison of four methods by varying $N$ from 1 to 5.
	}
\label{fig:multiway}
\end{figure}

\smallbreakparagraph{Weakly-supervised \ourmethod.}
The \ourmethod framework is versatile across the level of supervision: weak labels (class tags) or strong labels (segmentation masks).
Assuming weak labels are available but strong labels are not, \ournet is trainable with the classification learning objective of \ourmethod (Eq.~\ref{eq:loss_cls_final}) and its results are presented as $\text{\ournet}_{\text{w}}$ in \tableref{table:ipa}.
$\text{\ournet}_{\text{w}}$ performs on par with \ournet in terms of classification ER (82.0\% \textit{vs.}~84.9\% on 1-way 1-shot), but performs ineffectively on the segmentation task (15.0\% \textit{vs.}~52.3\% on 1-way 1-shot).
The result implies that the class tag labels are sufficient for a model to recognize the class occurrences, but are weak to endorse model's precise spatial recognition ability.

\smallbreakparagraph{Multi-class scalability of \ourtask.}
In addition, \ourtask is extensible to a multi-class problem with arbitrary numbers of classes, while \fsstask is not as flexible as \ourtask in the wild.
Figure~\ref{fig:multiway} compares the \ourtask performances of four methods by varying the $N$-way classes from one to five, where the other experimental setup follows the same one as in \tableref{table:ipa}.
Our \ournet shows consistently better performances than other methods on \ourtask in varying number of classes.

\begin{figure}[t!]
\includegraphics[width=0.33\linewidth]{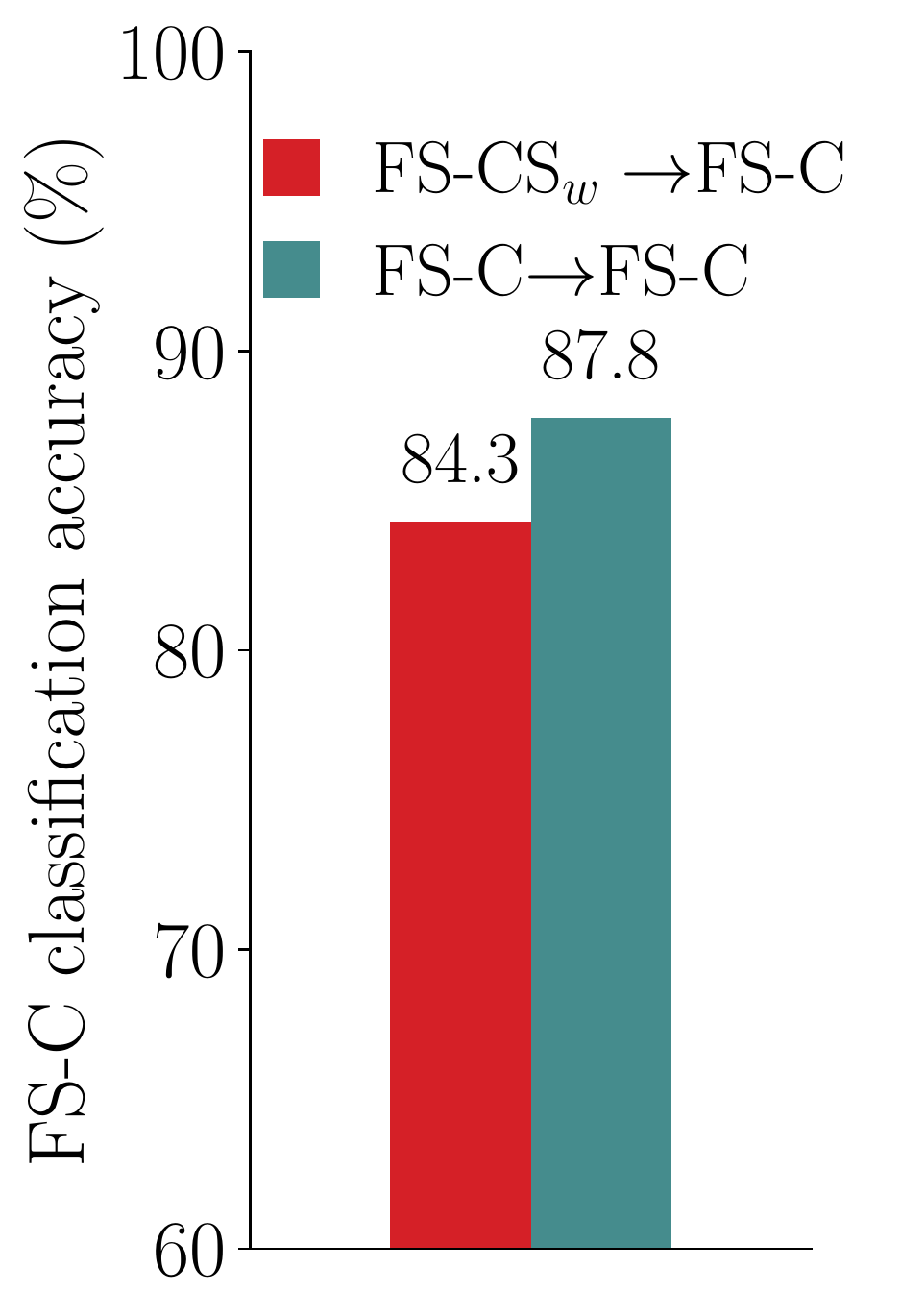}  \includegraphics[width=0.325\linewidth]{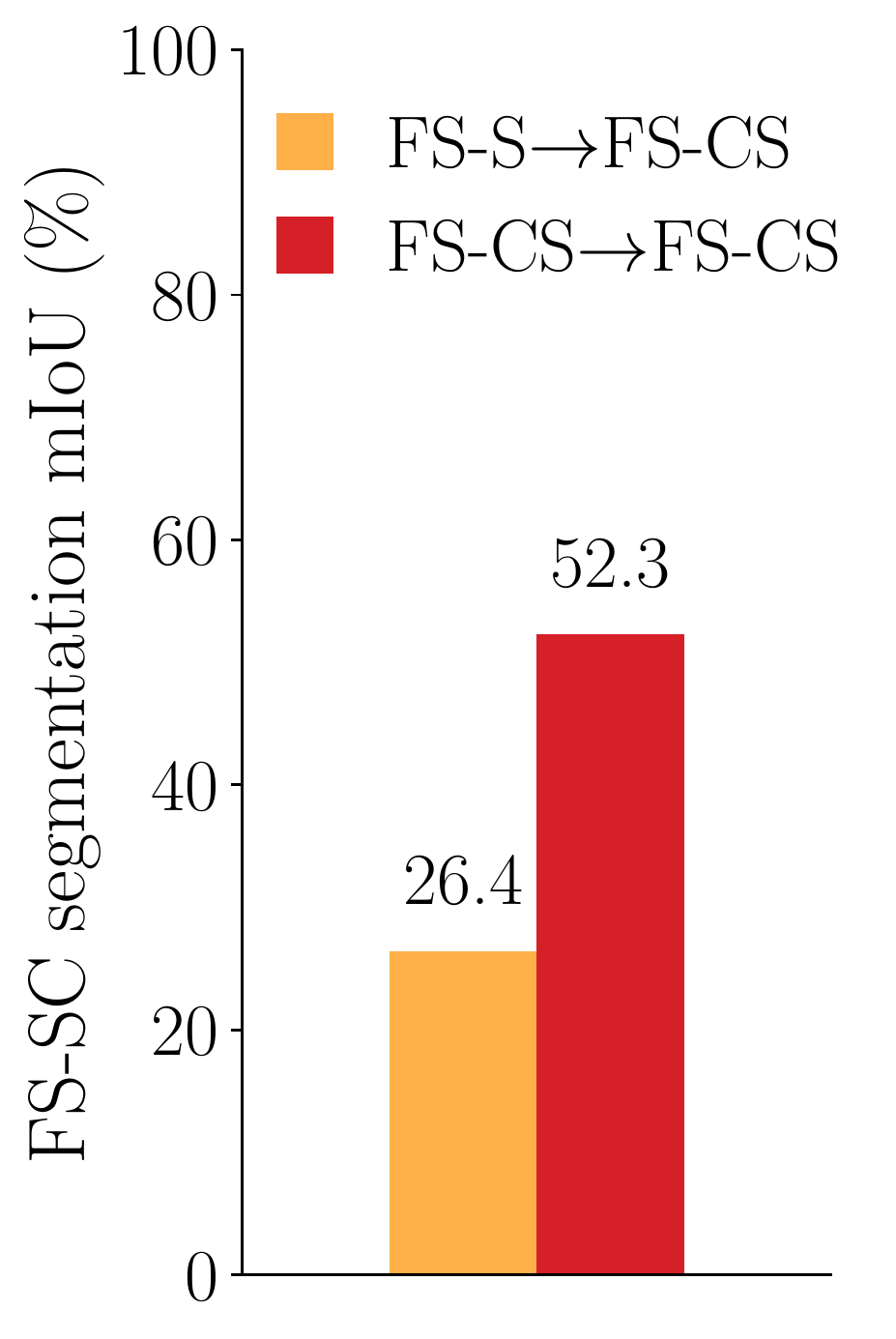} 
\includegraphics[width=0.32\linewidth]{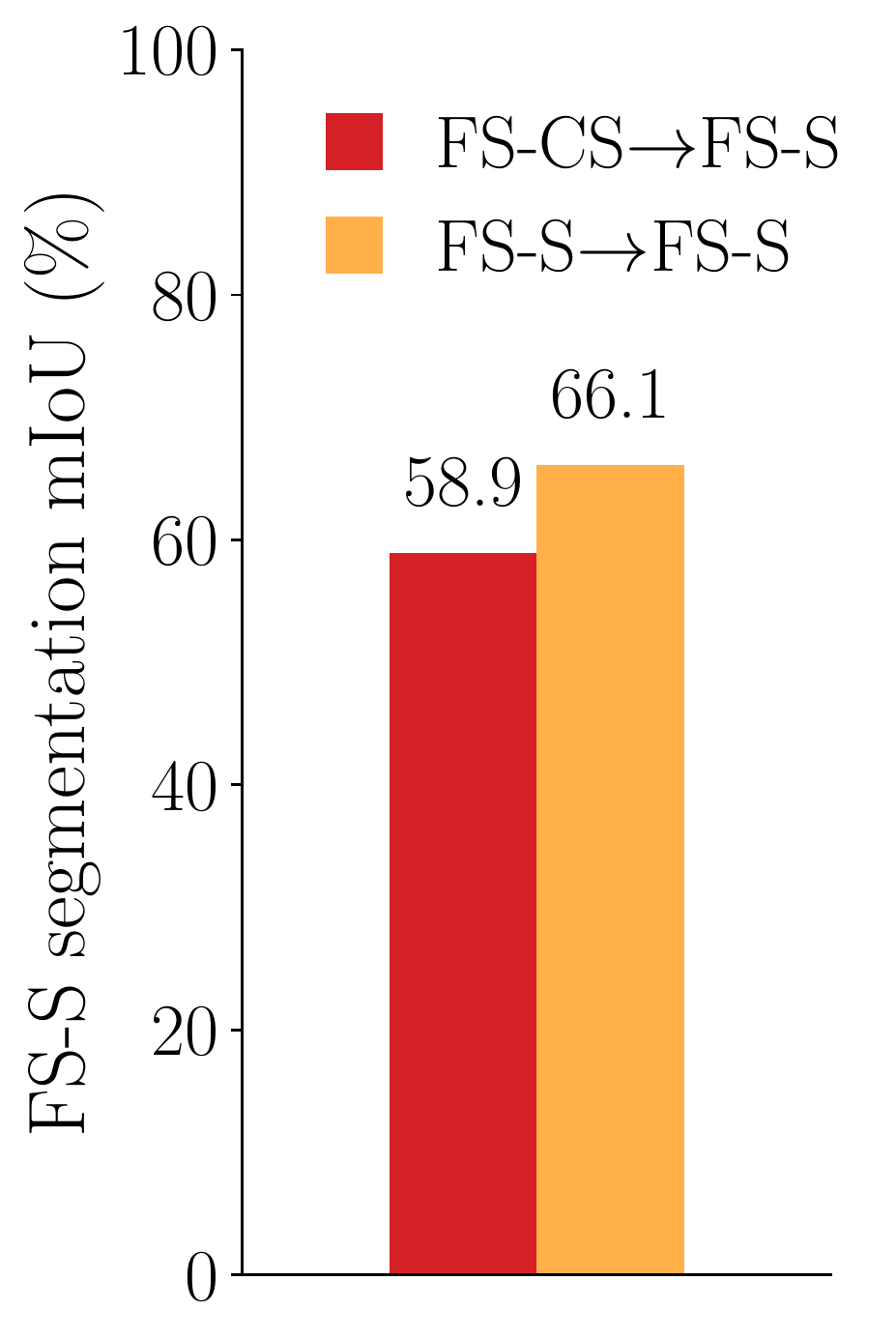} 
\hfill
\begin{minipage}[t]{.475\textwidth}
\vspace{-5mm}
\begin{minipage}[t]{.33\textwidth}
\centering \; \; \; \footnotesize{(a)} \\
\; \; \; \: \footnotesize{$\{\text{FS-CS}_{\text{w}}, \text{FS-C} \}$}\\
\; \; \; \footnotesize{$\rightarrow$FS-C}
\end{minipage}\hfill 
\begin{minipage}[t]{.33\textwidth}
\centering \; \; \; \: \footnotesize{(b)} \\
\; \; \; \; \: \footnotesize{\{FS-S, FS-CS\}}\\
\; \; \; \; \footnotesize{$\rightarrow$FS-CS}
\end{minipage}\hfill 
\begin{minipage}[t]{.33\textwidth}
\centering \; \; \; \: \footnotesize{(c)} \\
\; \; \; \; \: \footnotesize{\{FS-CS, FS-S\}}\\
\; \; \; \; \footnotesize{$\rightarrow$FS-S}
\end{minipage} 
\end{minipage}
\vspace{-2mm}
\caption{
Results of task transfer.
A $\rightarrow$ B denotes a model trained on task A and evaluated on task B.
$\text{\ourtask}_{\text{w}}$ denotes \ourtask with weak labels. 
\textbf{(a)}: Exclusive 2-way 1-shot classification accuracy of \fsctask or $\text{\ourtask}_{\text{w}}$ learners on \fsctask.
\textbf{(b)}: 1-way 1-shot segmentation mIoU of \fsstask or \ourtask learners on \ourtask.
\textbf{(c)}: 1-way 1-shot segmentation mIoU of \fsstask or \ourtask learners on \fsstask.
}
\label{fig:task_transfer}
\end{figure}

\smallbreakparagraph{Robustness of \ourtask against task transfer.}
We evaluate the transferability between \ourtask, \fsctask, and \fsstask by training a model on one task and evaluating it on the other task.
The results are compared in \figref{fig:task_transfer} in which `$\text{\fsstask} \rightarrow$ \ourtask' represents the result where the model trained on the $\text{\fsstask}$ task (with the guarantee of support class presence) is evaluated on the \ourtask setup.
To construct training and validation splits for \fsctask or \fsstask, we sample episodes that satisfy the constraint of support class occurrences~\footnote{We sample 2-way 1-shot episodes having a single positive class for training on \fsctask or evaluating on \fsctask.
We collect 1-way 1-shot episodes sampled from the same class for training on \fsstask or evaluating on \fsstask.}. 
For training \fsctask models, we use the class tag supervision only. 
All the other settings are fixed the same, \eg, we use \ournet with ResNet50 and Pascal-$i^5$.

The results show that \ourtask learners, \ie, models trained on \ourtask, are transferable to the two conventional few-shot learning tasks and yet overcome their shortcomings.
The transferability between few-shot classification tasks, \ie, \fsctask and $\text{\ourtask}_{\text{w}}$, is presented in \figref{fig:task_transfer}(a).
On this setup, the $\text{\ourtask}_{\text{w}}$ learner is evaluated by predicting a higher class response between the two classes, although it is trained using the multi-label classification objective.
The \ourtask learner closely competes with the \fsctask learner on \fsctask in terms of classification accuracy.
In contrast, the task transfer between segmentation tasks, \fsstask and \ourtask, results in asymmetric outcomes as shown in \figref{fig:task_transfer}(b)~and~(c).
The \ourtask learner shows relatively small performance drop on \fsstask, however, the \fsstask learner suffers a severe performance drop on \ourtask.
Qualitative examples in \figref{fig:teaser} demonstrate that the \fsstask learner predicts a vast number of false-positive pixels and results in poor performances.
In contrast, the \ourtask learner successfully distinguishes the region of interest by analyzing the semantic relevance of the query objects between the support set.

\begin{table*}[t!]
    \centering
    \small
    \scalebox{1.0}{
        \begin{tabular}{llcccccc||ccccccr}
            \toprule
             & & \multicolumn{6}{c}{1-way 1-shot} & \multicolumn{6}{c}{1-way 5-shot} & \small{\# learn.} \\
             \cmidrule(lr){3-8}\cmidrule(lr){9-14}
             \multicolumn{2}{c}{method} & $5^{0}$ & $5^{1}$ & $5^{2}$ & $5^{3}$ & mIoU & FBIoU & $5^{0}$ & $5^{1}$ & $5^{2}$ & $5^{3}$ & mIoU & FBIoU & \small{params.}\\
             \midrule
             \multirow{9}{*}{R50} 
             & CANet  \cite{zhang2019canet} & 52.5 & 65.9 & 51.3 & 51.9 &  55.4 & 66.2 & 55.5 & 67.8 & 51.9 & 53.2 & 57.1 & 69.6 & - \\ 
             & PPNet \cite{liu2020ppnet} & 47.8 & 58.8 & 53.8 & 45.6 & 51.5 & 69.2 & 58.4 & 67.8 & 64.9 & 56.7 & 62.0 & 75.8 & 23.5 M \\ 
             & PFENet \cite{tian2020pfenet} & 61.7 & 69.5 & 55.4 & 56.3 & 60.8 & 73.3 & 63.1 & 70.7 & 55.8 & 57.9 & 61.9 & 73.9 & 31.5 M \\ 
             & SAGNN \cite{xie2021scale} & 64.7 & 69.6 & 57.0 & 57.2 & 62.1 & 73.2 & 64.9 & 70.0 & 57.0 & 59.3 & 62.8 & 73.3 & - \\  
             & MMNet \cite{wu2021learning} & 62.7 & 70.2 & 57.3 & 57.0 & 61.8 & - & 62.2 & 71.5 & 57.5 & 62.4 & 63.4 & - & 10.4 M \\ 
             & CMN \cite{xie2021few} & 64.3 & 70.0 & 57.4 & 59.4 & 62.8 & 72.3 & 65.8 & 70.4 & 57.6 & 60.8 & 63.7 & 72.8 & - \\  
             & MLC \cite{yang2021mining} & 59.2 & 71.2 & \textbf{65.6} & 52.5 & 62.1 & - & 63.5 & 71.6 & \textbf{71.2} & 58.1 & 66.1 & - & 8.7 M\\  
             & HSNet \cite{hsnet} & 64.3 & 70.7 & 60.3 & 60.5 & 64.0 & 76.7 & 70.3 & 73.2 & 67.4 & \textbf{67.1} & 69.5 & \textbf{80.6} & 2.6 M \\   
             & \ccol \ournet & \ccol \textbf{68.9} & \ccol \textbf{71.7} & \ccol 61.1 & \ccol \textbf{62.7} & \ccol \textbf{66.1} & \ccol \textbf{77.7} & \ccol \textbf{72.6} & \ccol \textbf{74.3} & \ccol 65.3 & \ccol \textbf{67.1} & \ccol \textbf{70.8} & \ccol 80.4 & \ccol \textbf{1.3 M} \\     
             \bottomrule
        \end{tabular}
    }
    \caption{\fsstask results on 1-way 1-shot and 1-way 5-shot setups on Pascal-5$^i$~\cite{shaban2017oslsm} using ResNet50~\cite{resnet} (R50).
    }
    \label{table:fpa}
    \vspace{-1mm}
\end{table*}

\begin{table}[t!]
    \centering
    \small
    \scalebox{0.86}{
        \begin{tabular}{llcc||ccr}
            \toprule
             & & \multicolumn{2}{c}{1-way 1-shot} & \multicolumn{2}{c}{1-way 5-shot} & \small{\# learn.} \\
             \cmidrule(lr){3-4}\cmidrule(lr){5-6}
             \multicolumn{2}{c}{method} & mIoU & FBIoU & mIoU & FBIoU & \small{params.}\\
             \midrule
             \multirow{7}{*}{R50} 
             & RPMM \cite{yang2020pmm}      & 30.6 & - & 35.5 & - & 38.6 M \\ 
             & RePRI \cite{malik2021repri}  & 34.0 & - & 42.1 & - & - \\ 
             & MMNet \cite{wu2021learning}  & 37.5 & - & 38.2 & - & 10.4 M \\ 
             & MLC \cite{yang2021mining}    & 33.9 & - & 40.6 & - & 8.7 M \\ 
             & CMN \cite{xie2021few}        & 39.3 & 61.7 & 43.1 & 63.3 & - \\ 
             & HSNet \cite{hsnet}           & 39.2 & 68.2 & 46.9 & 70.7 & 2.6 M \\   
             & \ccol \ournet                  & \ccol \textbf{42.2} & \ccol \textbf{68.8} & \ccol \textbf{47.9} & \ccol \textbf{71.6} & \ccol \textbf{1.3 M} \\      
             \bottomrule
        \end{tabular}
    }
    \vspace{-1mm}
    \caption{\fsstask results on 1-way 1-shot and 1-way 5-shot setups on COCO-20$^i$~\cite{nguyen2019fwb}.}
    \label{table:fco}
    \vspace{-3mm}
\end{table}

\begin{table}[t!]
    \centering
    \scalebox{0.9}{
        \begin{tabular}{lcc}
            \toprule
            method & ER & mIoU \\
            \midrule
            \texttt{(a)} global $\rightarrow$ local & 83.9 & 44.6 \\
            \texttt{(b)} w/o masked attention & 83.8 & 50.8 \\
            \texttt{(c)} w/o multi-layer fusion & 83.1 & 51.6 \\ 
            \ccol \ournet & \ccol 84.9 & \ccol 52.3 \\
            \bottomrule
        \end{tabular}
    } 
\vspace{-2mm}
    \captionof{table}{Ablation study of the \ourlayer layer on 1-way 1-shot on Pascal-5$^i$~\cite{shaban2017oslsm} using ResNet50~\cite{resnet}.
    \label{table:ablation_architecture}}
\vspace{-5mm}
\end{table}

\subsection{Comparison with recent \fsstask methods on \fsstask}
Tables~\ref{table:fpa} and \ref{table:fco} compare the results of the recent few-shot semantic segmentation methods and \ournet on the conventional \fsstask task.
All model performances in the tables are taken from corresponding papers, and the numbers of learnable parameters are either taken from papers or counted from their official sources of implementation.
For a fair comparison with each other, some methods that incorporate extra unlabeled images~\cite{yang2021mining, liu2020ppnet} are reported as their model performances measured in the absence of the extra data.
Note that \ournet in Tables~\ref{table:fpa} and \ref{table:fco} is trained and evaluated following the \fsstask setup, not the proposed \ourtask one.

The results verify that \ournet outperforms the existing methods including the most recent ones~\cite{wu2021learning, xie2021few, yang2021mining}.
Especially, the methods that cast few-shot segmentation as the task of correlation feature transform, ASNet and HSNet~\cite{hsnet}, outperform other visual feature transform methods, indicating that learning correlations is beneficial for both \ourtask and \fsstask.
Note that \ournet is the most lightweight among others as \ournet processes correlation features that have smaller channel dimensions, \eg, at most 128, than visual features, \eg, at most 2048 in ResNet50.

\subsection{Analyses on the model architecture}
We perform ablation studies on the model architecture to reveal the benefit of each component.
We replace the global self-attention in the \ournet layer with the local self-attention~\cite{lsa} to see the effect of the global self-attention~(\tableref{table:ablation_architecture}\texttt{a}).
The local self-attention variant is compatible with the global \ournet in terms of the classification exact ratio but degrades the segmentation mIoU significantly, signifying the importance of the learning the global context of feature correlations.
Next, we ablate the attention masking in \eqref{eq:masked_attn}, which verifies that the attention masking prior is effective~(\tableref{table:ablation_architecture}\texttt{b}).
Lastly, we replace the multi-layer fusion path with spatial average pooling over the support dimensions followed by element-wise addition~(\tableref{table:ablation_architecture}\texttt{c}), and the result indicates that it is crucial to fuse outputs from the multi-layer correlations to precisely estimate class occurrence and segmentation masks.

\section{Discussion}
We have introduced the integrative task of few-shot classification and segmentation (\ourtask) that generalizes two existing few-shot learning problems.
Our proposed integrative few-shot learning (\ourmethod) framework is shown to be effective on \ourtask, in addition, our proposed attentive squeeze network (\ournet) outperforms recent state-of-the-art methods on both \ourtask and \fsstask.
The \ourmethod design allows a model to learn either with weak or strong labels, that being said, 
learning our method with weak labels achieves low segmentation performances. 
This result opens a future direction of effectively boosting the segmentation performance leveraging weak labels in the absence of strong labels for \ourtask.

\smallbreakparagraph{Acknowledgements.}
This work was supported by Samsung Advanced Institute of Technology (SAIT) and also by Center for Applied Research in Artificial Intelligence (CARAI) grant funded by DAPA and ADD (UD190031RD). 

{\small
\bibliographystyle{ieee_fullname}
\bibliography{egbib}
}

\clearpage
\renewcommand{\theequation}{a.\arabic{equation}}
\renewcommand{\thetable}{a.\arabic{table}}
\renewcommand{\thefigure}{a.\arabic{figure}}
\renewcommand*{\thefootnote}{\arabic{footnote}}
\renewcommand\thesection{\Alph{section}}
\setcounter{section}{0}

\section{Supplementary Material}

\subsection{Detailed model architecture}
\begin{figure}[t!]
	\centering
	\small
    \includegraphics[width=\linewidth]{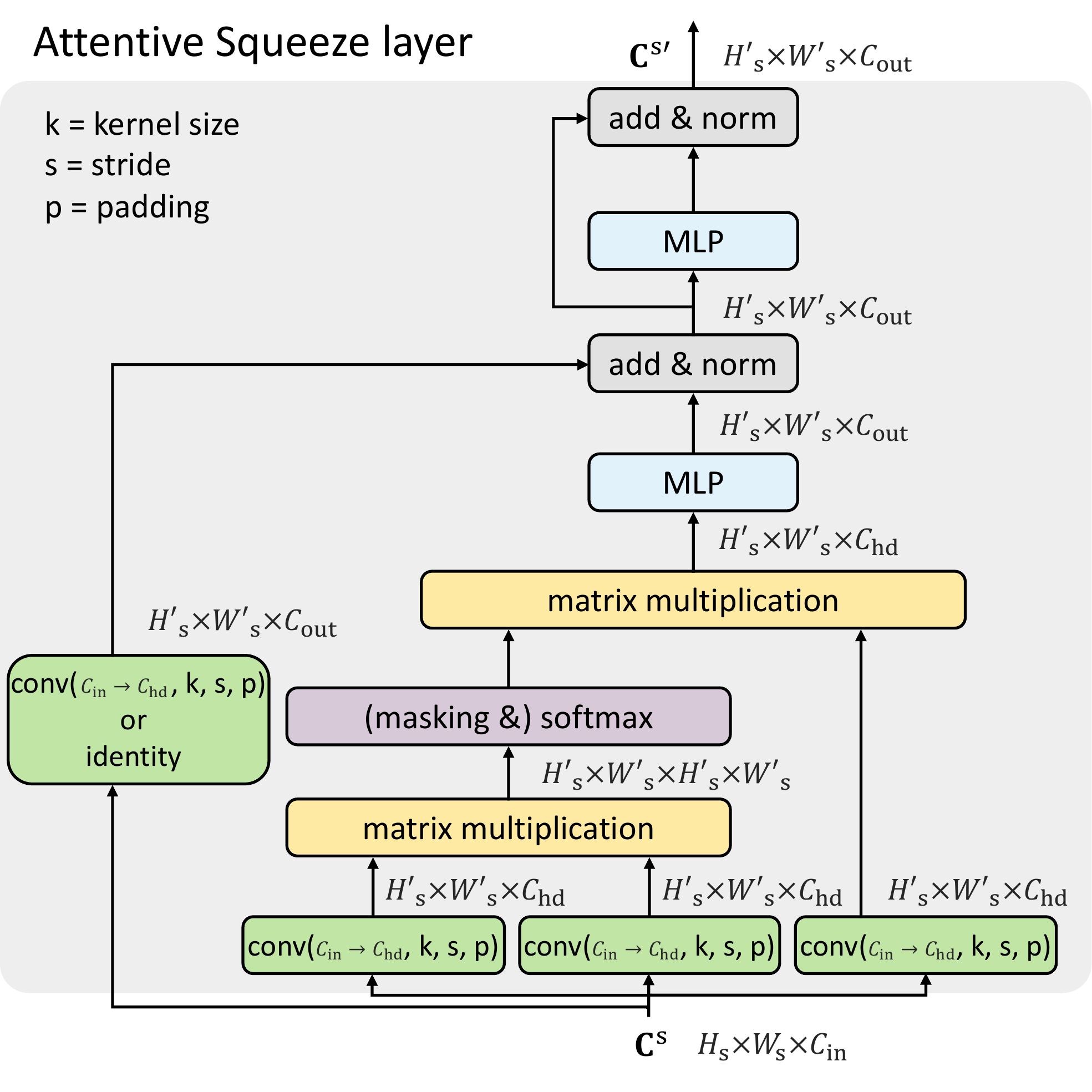}
    \vspace{-5mm}
	\caption{Illustration of the proposed attentive squeeze layer (Sec.~5.1. in the main paper).
	The shape of each output tensor is denoted next to arrows.
    }
\label{fig:aslayer}
\end{figure}

The comprehensive configuration of attentive squeeze network is summarized in \tableref{table:asnet}, and its building block, attentive squeeze layer, is depicted in \figref{fig:aslayer}.
The channel sizes of the input correlation $\{C_\text{in}^{(1)}, C_\text{in}^{(2)}, C_\text{in}^{(3)}\}$ corresponds to $\{4, 6, 3\}$, $\{4, 23, 3\}$, $\{3, 3, 1\}$ for ResNet50~\cite{resnet}, ResNet101, VGG-16~\cite{vgg}, respectively.

\subsection{Implementation details}
Our framework is implemented on PyTorch~\cite{pytorch} using the PyTorch Lightning~\cite{falcon2019pytorch} framework.
To reproduce the existing methods, we heavily borrow publicly available code bases.~\footnote{PANet~\cite{wang2019panet}: \url{https://github.com/kaixin96/PANet} \\ PFENet~\cite{tian2020pfenet}: \url{https://github.com/dvlab-research/PFENet} \\
HSNet~\cite{hsnet}: \url{https://github.com/juhongm999/hsnet}}
We set the officially provided hyper-parameters for each method while sharing generic techniques for all the methods, \eg, excluding images of small support objects for support sets or switching the role between the query and the support during training. 
NVIDIA GeForce RTX 2080 Ti GPUs or NVIDIA TITAN Xp GPUs are used in all experiments, where we train models using two GPUs on Pascal-$5^{i}$~\cite{shaban2017oslsm} while using four GPUs on COCO-$20^{i}$~\cite{nguyen2019fwb}.
Model training is halt either when it reaches the maximum $500_{\text{th}}$ epoch or when it starts to overfit.
We resize input images to $400 \times 400$ without any data augmentation strategies during both training and testing time for all methods.
For segmentation evaluation, we recover the two-channel output foreground map to its original image size by bilinear interpolation.
Pascal-$5^{i}$ and COCO-$20^{i}$ is derived from Pascal Visual Object Classes 2012~\cite{pascal} and Microsoft Common Object in Context 2014~\cite{coco}, respectively.
To construct episodes from datasets, we sample support sets such that one of the query classes is included in the support set by the probability of 0.5 to balance the ratio of background episodes across arbitrary benchmarks.

\newcommand{\plh}{%
  {\ooalign{$\phantom{0}$\cr\hidewidth$\scriptstyle\times$\cr}}%
}

\begin{table}[t!]
    \centering
    \setlength{\extrarowheight}{-2.0pt}
    \scalebox{0.70}{
        \begin{tabular}{ccc}
            \toprule
            $p = 1$ & $p = 2$ & $p = 3$ \\ 
            $\frac{H}{8}\plh\frac{H}{8}\plh\frac{H}{8}\plh\frac{H}{8}\plh C_\text{in}^{(1)}$ & $\frac{H}{16}\plh\frac{H}{16}\plh\frac{H}{16}\plh\frac{H}{16}\plh C_\text{in}^{(2)}$ & $\frac{H}{32}\plh\frac{H}{32}\plh\frac{H}{32}\plh\frac{H}{32}\plh C_\text{in}^{(3)}$ \\ 
            \midrule
            {[pool support dims. by half]} & & \\
            $\mathrm{AS}(C_\text{in}^{(1)}\rightarrow32, 5, 4, 2)$ & $\mathrm{AS}(C_\text{in}^{(2)}\rightarrow32, 5, 4, 2)$ & $\mathrm{AS}(C_\text{in}^{(3)}\rightarrow32, 5, 4, 2)$ \\
            $\mathrm{AS}(32\rightarrow128, 5, 4, 2$) & $\mathrm{AS}(32\rightarrow128, 5, 4, 2$) & $\mathrm{AS}(32\rightarrow128, 3, 2, 1$) \\
            {[pool support dims.]} & & \\
            {[upsample query dims.]} & & \\
            \multicolumn{2}{c}{ {[element-wise addition]} } & \\
            \multicolumn{2}{c}{ $\mathrm{AS}(128\rightarrow128, 1, 1, 0)$ } & \\
            \multicolumn{2}{c}{ $\mathrm{AS}(128\rightarrow128, 2, 1, 0)$ } & \\
            \multicolumn{2}{c}{ {[upsample query dims.]} } \\
            & \multicolumn{2}{c}{ {[element-wise addition]} } \\
            & \multicolumn{2}{c}{ $\mathrm{AS}(128\rightarrow128, 1, 1, 0)$ } \\
            & \multicolumn{2}{c}{ $\mathrm{AS}(128\rightarrow128, 2, 1, 0)$ } \\
            & \multicolumn{2}{c}{ $\mathrm{conv}(128\rightarrow128, 3, 1, 1)$ } \\
            & \multicolumn{2}{c}{ $\mathrm{ReLU}$ } \\
            & \multicolumn{2}{c}{ $\mathrm{conv}(128\rightarrow64, 3, 1, 1)$ } \\
            & \multicolumn{2}{c}{ $\mathrm{ReLU}$ } \\
            & \multicolumn{2}{c}{ {[upsample query dims.]} } \\
            & \multicolumn{2}{c}{ $\mathrm{conv}(64\rightarrow64, 3, 1, 1)$ } \\
            & \multicolumn{2}{c}{ $\mathrm{ReLU}$ } \\
            & \multicolumn{2}{c}{ $\mathrm{conv}(64\rightarrow2, 3, 1, 1)$ } \\
            & \multicolumn{2}{c}{ {[interpolate query dims. to the input size]} } \\
            \bottomrule
        \end{tabular}
    }  
    \caption{Comprehensive configuration of \ournet of which overview is illustrated in Fig.~2 in the main paper.
    The top of the table is the input of the model and the detailed architecture of the model below it.
    $\mathrm{AS}(C_{\text{in}} \rightarrow C_{\text{out}}, k, s, p)$ denotes an \ourlayer layer of the kernel size ($k$), stride ($s$), padding size ($p$) for the convolutional embedding with the input channel ($C_{\text{in}}$) and output channel~($C_{\text{out}}$). \label{table:asnet}
    }
\end{table}

\subsection{Further analyses}
In this subsection we provide supplementary analyses on the \ourmethod framework and \ournet.
All experimental results are obtained using ResNet50 on Pascal-$5^{i}$ and evaluated with 1-way 1-shot episodes unless specified otherwise.

\begin{figure}[t!]
	\centering
	\small
    \includegraphics[width=0.9\linewidth]{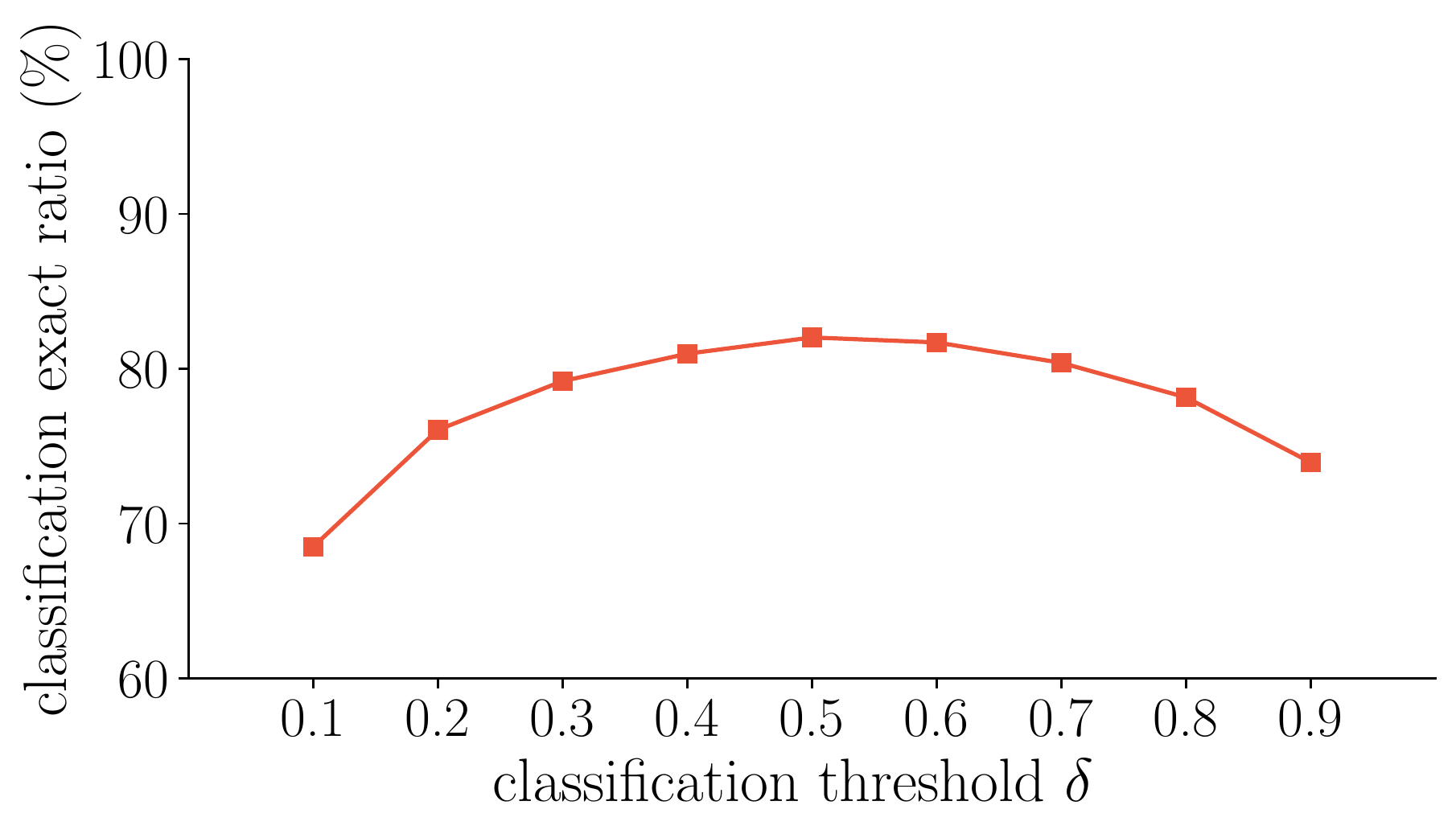}
	\caption{Classification threshold $\delta$ and its effects.
    }
\label{fig:cls_thr}
\vspace{-3mm}
\end{figure}

\smallbreakparagraph{The classification occurrence threshold $\delta$.}
Equation~2 in the main paper describes the process of detecting object classes on the shared foreground map by thresholding the highest foreground probability response on each foreground map.
As the foreground probability is bounded from 0 to 1, we set the threshold $\delta=0.5$ for simplicity.
A high threshold value makes a classifier reject insufficient probabilities as class presences.
Figure~\ref{fig:cls_thr} shows the classification 0/1 exact ratios by varying the threshold, which reaches the highest classification performance around $\delta=0.5$ and $0.6$.
Fine-tuning the threshold for the best classification performance is not the focus of this work, thus we opt for the most straightforward threshold $\delta=0.5$ for all experiments.

\begin{figure}[t!]
	\centering
	\small
	\includegraphics[width=0.99\linewidth]{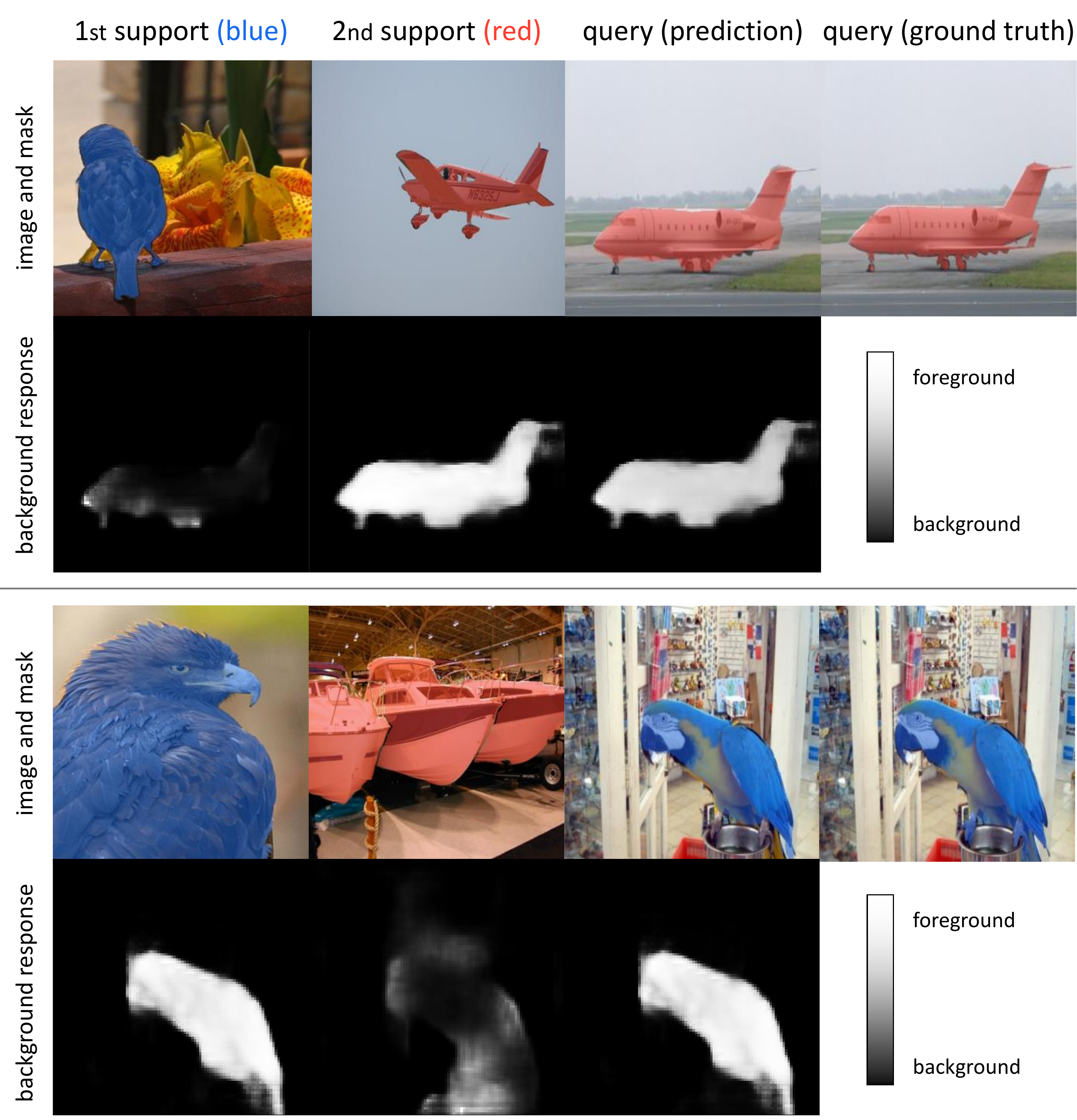}
	\caption{Visualization of background map for each support class and the merged background map $\bY_{\text{bg}}$ for the query.
	High background response is illustrated in black.
	}
\label{fig:bg}
\vspace{-3mm}
\end{figure}

\smallbreakparagraph{Visualization of $\bY_{\text{bg}}$.}
Figure~\ref{fig:bg} visually demonstrates the background merging step of \ourmethod in Eq.~(3) in the main paper.
The background maps are taken from the 2-way 1-shot episodes.
The background response of the negative class is relatively even, \ie, the majority of pixels are estimated as background, whereas the background response of the positive class highly contributes to the merged background map.

\begin{table*}[t!]
    \centering
    \small
    \setlength{\tabcolsep}{5pt}
    \scalebox{0.85}{
        \begin{tabular}{lccccc|ccccc||ccccc|cccccr}
            \toprule
             & \multicolumn{10}{c}{1-way 1-shot} & \multicolumn{10}{c}{2-way 1-shot} \\
             \cmidrule(lr){2-11}\cmidrule(lr){12-21}
             & \multicolumn{5}{c}{classification 0/1 exact ratio (\%)} & \multicolumn{5}{c}{segmentation mIoU (\%)} & \multicolumn{5}{c}{classification 0/1 exact ratio (\%)} & \multicolumn{5}{c}{segmentation mIoU (\%)} \\
             \cmidrule(lr){2-6}\cmidrule(lr){7-11}\cmidrule(lr){12-16}\cmidrule(lr){17-21}
             method & $5^{0}$ & $5^{1}$ & $5^{2}$ & $5^{3}$ & avg. & $5^{0}$ & $5^{1}$ & $5^{2}$ & $5^{3}$ & avg. & $5^{0}$ & $5^{1}$ & $5^{2}$ & $5^{3}$ & avg. & $5^{0}$ & $5^{1}$ & $5^{2}$ & $5^{3}$ & avg. \\
             \midrule
             \ournet ($\mathcal{L}_{\textcls}$)                          & 86.4 & 86.3 & 70.9 & 84.5 & 82.0 & 10.8 & 20.2 & 13.1 & 16.1 & 15.0 & \textbf{71.6} & 72.4 & 46.4 & 68.0 & 64.6 & 11.4 & 20.8 & 12.5 & 15.9 & 15.1  \\
             \ournet ($\mathcal{L}_{\textseg}$)                          & 84.9 & \textbf{89.6} & \textbf{79.0} & 86.2 & \textbf{84.9} & \textbf{51.7} & \textbf{61.5} & \textbf{43.3} & 52.8 & \textbf{52.3} & 68.5 & \textbf{76.2} & \textbf{58.6} & 70.0 & \textbf{68.3} & \textbf{48.5} & \textbf{58.3} & \textbf{36.3} & 48.3 & \textbf{47.8}  \\
             \ournet ($\mathcal{L}_{\textcls} + \mathcal{L}_{\textseg}$) & \textbf{86.9} & 87.4 & 75.8 & \textbf{88.7} & 84.7 & 51.6 & 61.2 & 42.4 & \textbf{53.2} & 52.1 & 70.1 & 72.4 & 54.8 & \textbf{74.8} & 68.0 & 48.1 & 57.1 & 36.0 & \textbf{50.1} & \textbf{47.8} \\
             \bottomrule
        \end{tabular}
    }
    \vspace{-2mm}
    \caption{
    \ourtask results of \ournet trained with \ourmethod objectives.
    $\mathcal{L}_{\textcls}$, $\mathcal{L}_{\textseg}$, and $\mathcal{L}_{\textcls} + \mathcal{L}_{\textseg}$ corresponds to \ourmethod learning objectives given classification tags, segmentation annotations, or both, respectively.
}
    \label{table:clsplussegloss}
    \vspace{-2mm}
\end{table*}

\begin{table*}[t!]
    \centering
    \small
    \setlength{\tabcolsep}{5pt}
    \scalebox{0.86}{
        \begin{tabular}{lccccc|ccccc||ccccc|cccccr}
            \toprule
             & \multicolumn{10}{c}{1-way 1-shot} & \multicolumn{10}{c}{2-way 1-shot} \\
             \cmidrule(lr){2-11}\cmidrule(lr){12-21}
             & \multicolumn{5}{c}{classification 0/1 exact ratio (\%)} & \multicolumn{5}{c}{segmentation mIoU (\%)} & \multicolumn{5}{c}{classification 0/1 exact ratio (\%)} & \multicolumn{5}{c}{segmentation mIoU (\%)} \\
             \cmidrule(lr){2-6}\cmidrule(lr){7-11}\cmidrule(lr){12-16}\cmidrule(lr){17-21}
             method & $5^{0}$ & $5^{1}$ & $5^{2}$ & $5^{3}$ & avg. & $5^{0}$ & $5^{1}$ & $5^{2}$ & $5^{3}$ & avg. & $5^{0}$ & $5^{1}$ & $5^{2}$ & $5^{3}$ & avg. & $5^{0}$ & $5^{1}$ & $5^{2}$ & $5^{3}$ & avg. \\
             \midrule
             PANet \cite{wang2019panet}     & 80.8 & 76.6 & 74.4 & 75.5 & 76.8 & 33.6 & 48.6 & 32.3 & 37.6 & 38.0 & 72.4 & 64.5 & 53.4 & 64.7 & 63.8 & 37.4 & 49.1 & 33.1 & 39.7 & 39.8 \\
             PFENet \cite{tian2020pfenet}   & 68.4 & 83.0 & 65.8 & 75.2 & 73.1 & 37.7 & 55.3 & 34.5 & 44.8 & 43.1 & 25.9 & 56.2 & 44.6 & 38.8 & 41.4 & 31.2 & 47.2 & 28.9 & 33.5 & 35.2 \\
             HSNet \cite{hsnet}             & 86.6 & 86.6 & 75.7 & 86.0 & 83.7 & 49.0 & 60.6 & 42.5 & 52.3 & 51.1 & 74.6 & 74.4 & 55.6 & 70.8 & 68.9 & 40.9 & 52.0 & 36.4 & 47.8 & 44.3 \\
             \ccol \ournet & \ccol \textbf{87.2} & \ccol \textbf{88.1} & \ccol \textbf{77.2} & \ccol \textbf{87.2} & \ccol \textbf{84.9} & \ccol \textbf{53.5} & \ccol \textbf{62.0} & \ccol \textbf{43.9} & \ccol \textbf{55.1} & \ccol \textbf{53.6} & \ccol \textbf{73.1} & \ccol \textbf{76.8} & \ccol \textbf{56.7} & \ccol \textbf{74.7} & \ccol \textbf{70.3} & \ccol \textbf{49.5} & \ccol \textbf{56.3} & \ccol \textbf{40.0} & \ccol \textbf{50.0} & \ccol \textbf{48.9} \\
             \bottomrule
        \end{tabular}
    }
    \vspace{-2mm}
    \caption{
    \ourtask results on Pascal-5$^i$ using ResNet101.
}
    \label{table:ipa101}
    \vspace{-3mm}
\end{table*}

\smallbreakparagraph{\ourmethod with weak labels, strong labels, and both.}
\tableref{table:clsplussegloss} compares \ourtask performances of three ASNets each of which trained with the classification loss (Eq.~(6) in the main paper), the segmentation loss (Eq.~(7) in the main paper), or both.
The loss is chosen upon the level of supervisions on support sets; classification tags (weak labels) or segmentation annotations (strong labels).
We observe that neither the classification nor segmentation performances deviate significantly between $\mathcal{L}_{\textseg}$ and $\mathcal{L}_{\textcls} + \mathcal{L}_{\textseg}$;
their performances are not even 0.3\%p different.
As a segmentation annotation is a dense form of classification tags, thus the classification loss influences insignificantly when the segmentation loss is used for training.
We thus choose to use the segmentation loss exclusively in the presence of segmentation annotations.

\subsection{Additional results}
Here we provide several extra experimental results that are omitted in the main paper due to the lack of space.
The contents include results using other backbone networks, another evaluation metric, and $K$ shots where $K > 1$.

\smallbreakparagraph{\ourmethod on \ourtask using ResNet101.}
We include the \ourtask results of the \ourmethod framework on Pascal-$5^{i}$ using ResNet101~\cite{resnet} in \tableref{table:ipa101}, which is missing in the main paper due to the page limit.  
All other experimental setups are matched with those of Table~1 in the main paper except for the backbone network.
\ournet also shows greater performances than the previous methods on both classification and segmentation tasks with another backbone.

\begin{table}[t!]
    \centering
    \small
    \setlength{\tabcolsep}{4.3pt}
    \scalebox{0.80}{
        \begin{tabular}{lccccc||ccccc}
            \toprule
             & \multicolumn{10}{c}{2-way 1-shot} \\
             \cmidrule(lr){2-6}\cmidrule(lr){7-11}
             & \multicolumn{5}{c}{classification 0/1 exact ratio (\%)} & \multicolumn{5}{c}{classification accuracy (\%)} \\
             \cmidrule(lr){2-6}\cmidrule(lr){7-11} 
             method & $5^{0}$ & $5^{1}$ & $5^{2}$ & $5^{3}$ & avg. & $5^{0}$ & $5^{1}$ & $5^{2}$ & $5^{3}$ & avg. \\
             \midrule
             PANet \cite{wang2019panet}     & 56.2 & 47.5 & 44.6 & 55.4 & 50.9 & 74.9 & 70.2 & 67.8 & 74.8 & 71.9 \\
             PFENet \cite{tian2020pfenet}   & 22.5 & 61.7 & 40.3 & 39.5 & 41.0 & 64.1 & 79.5 & 66.4 & 66.1 & 69.0 \\
             HSNet \cite{hsnet}             & 68.0 & 73.2 & 57.0 & \textbf{70.9} & 67.3 & 82.4 & 85.6 & 76.0 & \textbf{84.5} & 82.1 \\
             \ccol $\text{\ournet}_{\text{w}}$ & \ccol \textbf{71.6} & \ccol 72.1 & \ccol 46.4 & \ccol 68.0 & \ccol 64.6 & \ccol \textbf{84.9} & \ccol 85.4 & \ccol 69.2 & \ccol 82.2 & \ccol 80.4  \\
             \ccol \ournet & \ccol 68.5 & \ccol \textbf{76.2} & \ccol \textbf{58.6} & \ccol 70.0 & \ccol \textbf{68.3} & \ccol 82.9 & \ccol \textbf{87.5} & \ccol \textbf{76.7} & \ccol 84.0 & \ccol \textbf{82.8}  \\
             \bottomrule
        \end{tabular}
    }
    \vspace{-2mm}
    \caption{
    \ourtask classification accuracy (\%) and 0/1 exact ratio (\%) on Pascal-5$^i$ using ResNet50.
}
    \label{table:ipa50_acc}
    \vspace{-2mm}
\end{table}

\smallbreakparagraph{\ourtask classification metrics: 0/1 exact ratio and accuracy.}
\tableref{table:ipa50_acc} presents the results of two classification evaluation metrics of \ourtask: 0/1 exact ratio~\cite{durand2019learning} and classification accuracy.
The classification accuracy metric takes the average of correct predictions for each class for each query, while 0/1 exact ratio measures the binary correctness for all classes for each query, thus being stricter than the accuracy; the exact formulations are in Sec.~6.1. of the main paper.
\ournet shows higher classification performance in both classification metrics than others.

\begin{table*}[t!]
    \centering
    \small
    \setlength{\tabcolsep}{5pt}
    \scalebox{0.80}{
        \begin{tabular}{lccccc|ccccc||ccccc|cccccr}
            \toprule
             & \multicolumn{10}{c}{1-way 5-shot} & \multicolumn{10}{c}{2-way 5-shot} \\
             \cmidrule(lr){2-11}\cmidrule(lr){12-21}
             & \multicolumn{5}{c}{classification 0/1 exact ratio (\%)} & \multicolumn{5}{c}{segmentation mIoU (\%)} & \multicolumn{5}{c}{classification 0/1 exact ratio (\%)} & \multicolumn{5}{c}{segmentation mIoU (\%)} \\
             \cmidrule(lr){2-6}\cmidrule(lr){7-11}\cmidrule(lr){12-16}\cmidrule(lr){17-21}
             method & $5^{0}$ & $5^{1}$ & $5^{2}$ & $5^{3}$ & avg. & $5^{0}$ & $5^{1}$ & $5^{2}$ & $5^{3}$ & avg. & $5^{0}$ & $5^{1}$ & $5^{2}$ & $5^{3}$ & avg. & $5^{0}$ & $5^{1}$ & $5^{2}$ & $5^{3}$ & avg. \\
             \midrule
             PANet \cite{wang2019panet}     & 72.5 & 70.2 & 70.7 & 74.6 & 72.0 & 45.6 & 56.2 & 44.6 & 49.2 & 48.9 & 61.1 & 46.8 & 44.0 & 66.2 & 54.5 & 46.2 & 57.4 & 46.7 & 47.6 & 49.5 \\
             PFENet \cite{tian2020pfenet}   & 70.9 & 84.5 & 67.1 & 80.4 & 75.7 & 42.8 & 56.3 & 36.2 & 47.3 & 45.7 & 22.3 & 63.2 & 42.5 & 40.6 & 42.2 & 35.9 & 50.5 & 33.3 & 35.4 & 38.8 \\
             HSNet \cite{hsnet}             & \textbf{91.1} & 88.1 & 82.0 & 90.7 & 88.0 & 56.2 & 61.3 & 40.2 & 54.2 & 53.0 & 79.7 & 81.0 & 65.0 & \textbf{81.0} & 76.7 & 42.5 & 58.9 & 32.0 & 44.1 & 44.4 \\
             \ccol \ournet & \ccol 90.5 & \ccol \textbf{90.4} & \ccol \textbf{82.3} & \ccol \textbf{91.8} & \ccol \textbf{88.8} & \ccol \textbf{59.2} & \ccol \textbf{63.5} & \ccol \textbf{41.2} & \ccol \textbf{58.7} & \ccol \textbf{55.7} & \ccol \textbf{81.4} & \ccol \textbf{81.4} & \ccol \textbf{68.0} & \ccol 80.6 & \ccol \textbf{77.9} & \ccol \textbf{53.4} & \ccol \textbf{60.4} & \ccol \textbf{35.9} & \ccol \textbf{50.6} & \ccol \textbf{50.1} \\
             \bottomrule
        \end{tabular}
    }
    \vspace{-2mm}
    \caption{
    \ourtask results on 5-shot setups on Pascal-5$^i$ using ResNet50.
    }
    \label{table:ipa50_5shot}
    \vspace{-2mm}
\end{table*}

\begin{table*}[t!]
    \centering
    \small
    \setlength{\tabcolsep}{5pt}
    \scalebox{0.80}{
        \begin{tabular}{lccccc|ccccc||ccccc|cccccr}
            \toprule
             & \multicolumn{10}{c}{1-way 5-shot} & \multicolumn{10}{c}{2-way 5-shot} \\
             \cmidrule(lr){2-11}\cmidrule(lr){12-21}
             & \multicolumn{5}{c}{classification 0/1 exact ratio (\%)} & \multicolumn{5}{c}{segmentation mIoU (\%)} & \multicolumn{5}{c}{classification 0/1 exact ratio (\%)} & \multicolumn{5}{c}{segmentation mIoU (\%)} \\
             \cmidrule(lr){2-6}\cmidrule(lr){7-11}\cmidrule(lr){12-16}\cmidrule(lr){17-21}
             method & $5^{0}$ & $5^{1}$ & $5^{2}$ & $5^{3}$ & avg. & $5^{0}$ & $5^{1}$ & $5^{2}$ & $5^{3}$ & avg. & $5^{0}$ & $5^{1}$ & $5^{2}$ & $5^{3}$ & avg. & $5^{0}$ & $5^{1}$ & $5^{2}$ & $5^{3}$ & avg. \\
             \midrule
             PANet \cite{wang2019panet}     & 83.7 & 81.6 & 78.3 & 81.3 & 81.2 & 48.2 & 59.1 & \textbf{45.5} & 50.5 & 50.8 & 79.0 & 68.4 & 60.5 & 72.3 & 70.1 & 49.1 & 59.6 & \textbf{46.8} & 50.1 & \textbf{51.4} \\
             PFENet \cite{tian2020pfenet}   & 70.3 & 85.3 & 65.9 & 78.6 & 75.0 & 42.2 & 56.0 & 35.7 & 48.7 & 45.7 & 26.9 & 56.0 & 49.2 & 37.3 & 42.4 & 35.7 & 49.6 & 31.4 & 36.9 & 38.4 \\
             HSNet \cite{hsnet}             & 91.4 & 89.5 & 79.4 & 90.9 & 87.8 & 55.2 & 64.2 & 41.7 & 58.4 & 54.9 & \textbf{85.6} & 80.8 & 61.3 & 81.7 & 77.4 & 38.5 & 57.6 & 34.8 & 49.8 & 45.2 \\
             \ccol \ournet & \ccol \textbf{91.5} & \ccol \textbf{90.2} & \ccol \textbf{80.6} & \ccol \textbf{93.4} & \ccol \textbf{88.9} & \ccol \textbf{60.3} & \ccol \textbf{64.7} & \ccol 41.4 & \ccol \textbf{58.5} & \ccol \textbf{56.2} & \ccol 82.8 & \ccol \textbf{81.1} & \ccol \textbf{65.1} & \ccol \textbf{85.5} & \ccol \textbf{78.6} & \ccol \textbf{53.8} & \ccol \textbf{61.0} & \ccol 34.2 & \ccol \textbf{52.2} & \ccol 50.3 \\
             \bottomrule
        \end{tabular}
    }
    \vspace{-2mm}
    \caption{
    \ourtask results on 5-shot setups on Pascal-5$^i$ using ResNet101.
    }
    \label{table:ipa101_5shot}
    \vspace{-2mm}
\end{table*}

\smallbreakparagraph{\ourmethod on 5-shot \ourtask.}
Tables~\ref{table:ipa50_5shot} and \ref{table:ipa101_5shot} compares four different methods on the 1-way 5-shot and 2-way 5-shot \ourtask setups, which are missing in the main paper due to the page limit.
All other experimental setups are matched with those of Table~1 in the main paper except for the number of support samples for each class, \ie, varying $K$ shots.
\ournet also outperforms other methods on the multi-shot setups.

\begin{table*}[t!]
    \centering
    \small
    \scalebox{0.91}{
        \begin{tabular}{llcccccc||ccccccr}
            \toprule
             & & \multicolumn{6}{c}{1-way 1-shot} & \multicolumn{6}{c}{1-way 5-shot} & \small{\# learn.} \\
             \cmidrule(lr){3-8}\cmidrule(lr){9-14}
             \multicolumn{2}{c}{method} & $5^{0}$ & $5^{1}$ & $5^{2}$ & $5^{3}$ & mIoU & FBIoU & $5^{0}$ & $5^{1}$ & $5^{2}$ & $5^{3}$ & mIoU & FBIoU & \small{params.}\\
             \midrule
             \multirow{7}{*}{VGG-16} 
             & OSLSM \cite{shaban2017oslsm} & 33.6 & 55.3 & 40.9 & 33.5 & 40.8 &   -  & 35.9 & 58.1 & 42.7 & 39.1 & 43.9 &   -  & 276.7 M \\ 
             & PANet \cite{wang2019panet}   & 42.3 & 58.0 & 51.1 & 41.2 & 48.1 & 66.5 & 51.8 & 64.6 & 59.8 & 46.5 & 55.7 & 70.7 & 14.7 M \\ 
             & FWB \cite{nguyen2019fwb}     & 47.0 & 59.6 & 52.6 & 48.3 & 51.9 &   -  & 50.9 & 62.9 & 56.5 & 50.1 & 55.1 &   -  &   -  \\ 
             & RPMMs \cite{yang2020pmm}     & 47.1 & 65.8 & 50.6 & 48.5 & 53.0 &   -  & 50.0 & 66.5 & 51.9 & 47.6 & 54.0 &   -  &   -  \\ 
             & PFENet \cite{tian2020pfenet} & 56.9 & \textbf{68.2} & 54.4 & 52.4 & 58.0 & 72.0 & 59.0 & 69.1 & 54.8 & 52.9 & 59.0 & 72.3 & 10.4 M \\ 
             & HSNet \cite{hsnet} & 59.6 & 65.7 & \textbf{59.6} & 54.0 & 59.7 & \textbf{73.4} & 64.9 & 69.0 & \textbf{64.1} & 58.6 & 64.1 & \textbf{76.6} & 2.6 M \\ 
             & \ccol \ournet & \ccol \textbf{61.7} & \ccol 66.7 & \ccol 58.6 & \ccol \textbf{55.3} & \ccol \textbf{60.6} & \ccol 73.2 & \ccol \textbf{66.5} & \ccol \textbf{69.6} & \ccol 63.0 & \ccol \textbf{60.5} & \ccol \textbf{64.9} & \ccol 76.5 & \ccol \textbf{1.3 M} \\       
             \hline
             \multirow{7}{*}{R101} 
             & FWB \cite{nguyen2019fwb} & 51.3 & 64.5 & 56.7 & 52.2 & 56.2 & - & 54.8 & 67.4 & 62.2 & 55.3 & 59.9 & - & 43.0 M \\  
             & DAN \cite{wang2020dan} & 54.7 &  68.6 & 57.8 & 51.6 & 58.2 & 71.9 & 57.9 &  69.0 & 60.1 & 54.9 & 60.5 & 72.3 & - \\  
             & RePRI \cite{malik2021repri} &  59.6 & 68.6 & 62.2 & 47.2 & 59.4 & - & 66.2 & 71.4 & 67.0 & 57.7 & 65.6 & - & 65.7 M \\ 
             & PFENet \cite{tian2020pfenet} & 60.5 & 69.4 & 54.4 & 55.9 & 60.1 & 72.9 & 62.8 & 70.4 & 54.9 & 57.6 & 61.4 & 73.5 & 10.8 M\\  
             & MLC \cite{yang2021mining} & 60.8 & 71.3 & 61.5 & 56.9 & 62.6 & - & 65.8 & 74.9 & \textbf{71.4} & 63.1 & 68.8 & - & 27.7 M \\  
             & HSNet \cite{hsnet} & 67.3 & 72.3 & \textbf{62.0} & 63.1 & 66.2 & 77.6 & 71.8 & 74.4 & 67.0 & 68.3 & 70.4 & 80.6 & 2.6 M \\  
             & \ccol \ournet & \ccol \textbf{69.0} & \ccol \textbf{73.1} & \ccol \textbf{62.0} & \ccol \textbf{63.6} & \ccol \textbf{66.9} & \ccol \textbf{78.0} & \ccol \textbf{73.1} & \ccol \textbf{75.6} & \ccol 65.7 & \ccol \textbf{69.9} & \ccol \textbf{71.1} & \ccol \textbf{81.0} & \ccol \textbf{1.3 M} \\ 
             \bottomrule
        \end{tabular}
    }
    \caption{\fsstask results on 1-way 1-shot and 1-way 5-shot setups on PASCAL-5$^i$ using VGG-16~\cite{vgg} and ResNet101~\cite{resnet}.
    }
    \label{table:fpavgg}
    \vspace{-3mm}
\end{table*}

\begin{table}[t!]
    \centering
    \small
    \setlength{\tabcolsep}{4.3pt}
    \scalebox{0.78}{
        \begin{tabular}{lccccc||ccccc}
            \toprule
             & \multicolumn{10}{c}{$N$-way 1-shot}  \\
             \cmidrule(lr){2-11} 
             & \multicolumn{5}{c}{classification 0/1 exact ratio (\%)} & \multicolumn{5}{c}{segmentation mIoU (\%)} \\
             \cmidrule(lr){2-6}\cmidrule(lr){7-11} 
             method & 1 & 2 & 3 & 4 & 5 & 1 & 2 & 3 & 4 & 5 \\
             \midrule
             PANet \cite{wang2019panet}     & 69.0 & 50.9 & 39.3 & 29.1 & 22.2 & 36.2 & 37.2 & 37.1 & 36.6 & 35.3 \\
             PFENet \cite{tian2020pfenet}   & 74.6 & 41.0 & 24.9 & 14.5 &  7.9 & 43.0 & 35.3 & 30.8 & 27.6 & 24.9 \\
             HSNet \cite{hsnet}             & 82.7 & 67.3 & 52.5 & 45.2 & 36.8 & 49.7 & 43.5 & 39.8 & 38.1 & 36.2 \\
             \ccol \ournet & \ccol \textbf{84.9} & \ccol \textbf{68.3} & \ccol \textbf{55.8} & \ccol \textbf{46.8} & \ccol \textbf{37.3} & \ccol \textbf{52.3} & \ccol \textbf{47.8} & \ccol \textbf{45.4} & \ccol \textbf{44.5} & \ccol \textbf{42.4} \\
             \bottomrule
        \end{tabular}
    }
    \vspace{-2mm}
    \caption{
    Numerical results of Fig.~4 in the main paper:
    \ourtask performances on $N$-way 1-shot by varying $N$ from 1 to 5.
}
    \label{table:multiway}
    \vspace{-2mm}
\end{table}

\smallbreakparagraph{\ournet on \fsstask using VGG-16.}
\tableref{table:fpavgg} compares the recent state-of-the-art methods and \ournet on \fsstask using VGG-16~\cite{vgg}.
We train and evaluate \ournet with the \fsstask problem setup to fairly compare with the recent methods.
All the other experimental variables are detailed in Sec.~6.3. and Table~3 of the main paper.
\ournet consistently shows outstanding performances using the VGG-16 backbone network as observed in experimnets using ResNets.

\begin{figure*}[t!]
	\centering
	\small
	\includegraphics[width=\linewidth]{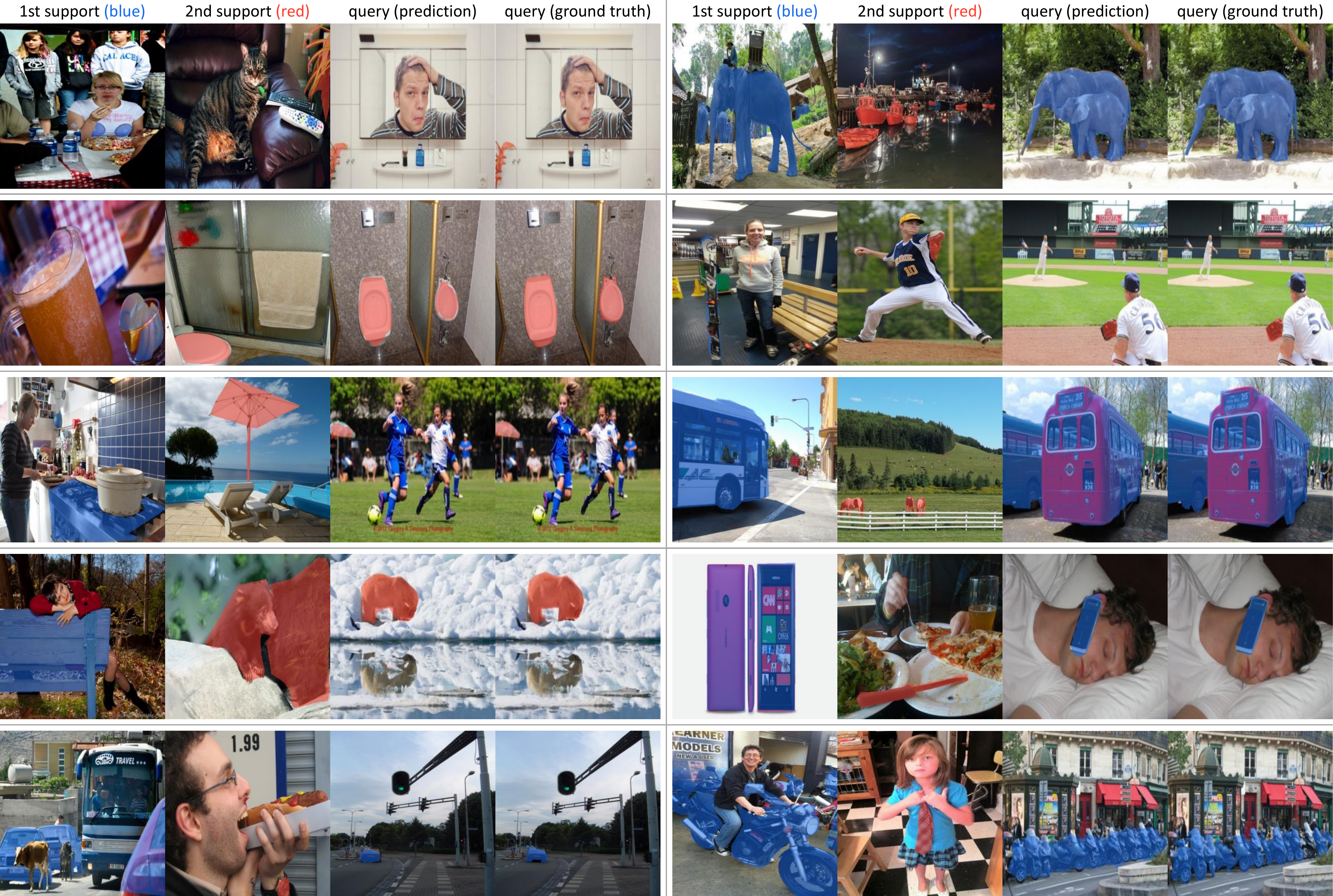}
	\caption{2-way 1-shot \ourtask segmentation prediction maps on the COCO-$20^{i}$ benchmark.}
\label{fig:2way_extra_coco}
\end{figure*}

\smallbreakparagraph{Qualitative results.}
We attach additional segmentation predictions of \ournet learned with the \ourmethod framework on the \ourtask task in \figref{fig:2way_extra_coco}.
We observe that \ournet successfully predicts segmentation maps at challenging scenarios in the wild such as a) segmenting tiny objects, b) segmenting non-salient objects, c) segmenting multiple objects, and d) segmenting a query given a small support object annotation.

\begin{figure*}[t!]
	\centering
	\small
	\includegraphics[width=0.80\linewidth]{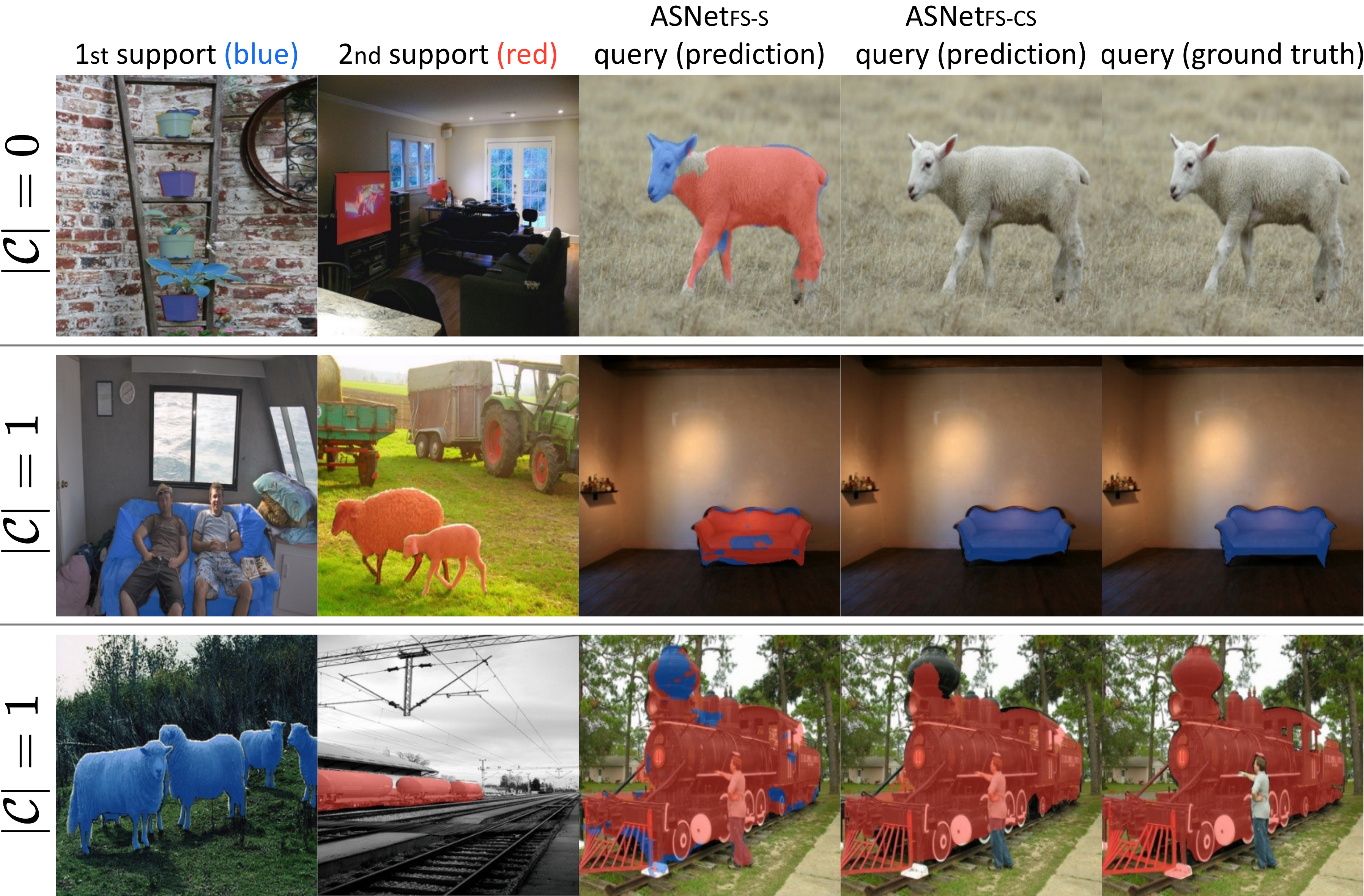}
	\caption{2-way 1-shot \ourtask segmentation prediction maps of $\text{ASNet}_{\text{\fsstask}}$ and $\text{ASNet}_{\text{\ourtask}}$.}
\label{fig:fssfcss}
\end{figure*}

\smallbreakparagraph{Qualitative results of $\text{ASNet}_{\text{\fsstask}}$.}
Figure~\ref{fig:fssfcss} visualizes typical failure cases of the $\text{ASNet}_{\text{\fsstask}}$ model in comparison with $\text{ASNet}_{\text{\ourtask}}$; these examples qualitatively show the severe performance drop of $\text{ASNet}_{\text{\fsstask}}$ on \ourtask, which is quantitatively presented in Fig.~5~(b) of the main paper.
Sharing the same architecture of \ournet, each model is trained on either \fsstask or \ourtask setup and evaluated on the 2-way 1-shot \ourtask setup.
The results demonstrate that $\text{ASNet}_{\text{\fsstask}}$ is unaware of object classes and gives foreground predictions on any existing objects, whereas $\text{ASNet}_{\text{\ourtask}}$ effectively distinguishes the object classes based on the support classes and produces clean and adequate segmentation maps.

\begin{table*}[t!]
    \centering
    \small
    \setlength{\tabcolsep}{5pt}
    \scalebox{0.86}{
        \begin{tabular}{lccccc|ccccc||ccccc|cccccr}
            \toprule
             & \multicolumn{10}{c}{1-way 1-shot} & \multicolumn{10}{c}{2-way 1-shot} \\
             \cmidrule(lr){2-11}\cmidrule(lr){12-21}
             & \multicolumn{5}{c}{classification 0/1 exact ratio (\%)} & \multicolumn{5}{c}{segmentation mIoU (\%)} & \multicolumn{5}{c}{classification 0/1 exact ratio (\%)} & \multicolumn{5}{c}{segmentation mIoU (\%)} \\
             \cmidrule(lr){2-6}\cmidrule(lr){7-11}\cmidrule(lr){12-16}\cmidrule(lr){17-21}
             method & $20^{0}$ & $20^{1}$ & $20^{2}$ & $20^{3}$ & avg. & $20^{0}$ & $20^{1}$ & $20^{2}$ & $20^{3}$ & avg. & $20^{0}$ & $20^{1}$ & $20^{2}$ & $20^{3}$ & avg. & $20^{0}$ & $20^{1}$ & $20^{2}$ & $20^{3}$ & avg. \\
             \midrule
             PANet \cite{wang2019panet}     & 64.3 & 66.5 & 68.0 & 67.9 & 66.7 & 25.5 & 24.7 & 25.7 & 24.7 & 25.2 & 42.5 & 49.9 & 53.6 & 47.8 & 48.5 & 24.9 & 25.0 & 23.3 & 21.4 & 23.6 \\
             PFENet \cite{tian2020pfenet}   & 70.7 & 70.6 & 71.2 & 72.9 & 71.4 & 30.6 & 34.8 & 29.4 & 32.6 & 31.9 & 35.6 & 34.3 & 43.1 & 32.8 & 36.5 & 23.3 & 23.8 & 20.2 & 23.1 & 22.6 \\
             HSNet \cite{hsnet}             & 74.7 & 77.2 & 78.5 & 77.6 & 77.0 & \textbf{36.2} & 34.3 & 32.9 & 34.0 & 34.3 & 57.7 & \textbf{62.4} & 67.1 & \textbf{62.6} & 62.5 & 28.9 & 29.6 & 30.3 & 29.3 & 29.5 \\
             \ccol \ournet & \ccol \textbf{76.2} & \ccol \textbf{78.8} & \ccol \textbf{79.2} & \ccol \textbf{80.2} & \ccol \textbf{78.6} & \ccol 35.7 & \ccol \textbf{36.8} & \ccol \textbf{35.3} & \ccol \textbf{35.6} & \ccol \textbf{35.8} & \ccol \textbf{59.5} & \ccol 61.5 & \ccol \textbf{68.8} & \ccol 62.4 & \ccol \textbf{63.1} & \ccol \textbf{29.8} & \ccol \textbf{33.0} & \ccol \textbf{33.4} & \ccol \textbf{30.4} & \ccol \textbf{31.6} \\

             \bottomrule
        \end{tabular}
    }
    \vspace{-2mm}
    \caption{
    Fold-wise \ourtask results on COCO-20$^i$ using ResNet50.
    The results correspond to the Table~2 in the main paper.
}
    \label{table:ico50_foldwise}
    \vspace{-4mm}
\end{table*}


\begin{table*}[t!]
    \centering
    \small
    \scalebox{0.97}{
        \begin{tabular}{llccccccccccccr}
            \toprule
             & & \multicolumn{6}{c}{1-way 1-shot} & \multicolumn{6}{c}{1-way 5-shot} & \small{\# learn.} \\
             \cmidrule(lr){3-8}\cmidrule(lr){9-14}
             \multicolumn{2}{c}{method} & $20^{0}$ & $20^{1}$ & $20^{2}$ & $20^{3}$ & mIoU & FBIoU & $20^{0}$ & $20^{1}$ & $20^{2}$ & $20^{3}$ & mIoU & FBIoU & \small{params.}\\
             \midrule
             \multirow{7}{*}{R50} 
             & RPMM \cite{yang2020pmm} & 29.5 & 36.8 & 28.9 & 27.0 & 30.6 & - & 33.8 & 42.0 & 33.0 & 33.3 & 35.5 & - & 38.6 M \\ 
             & RePRI \cite{malik2021repri} & 31.2 & 38.1 & 33.3 & 33.0 & 34.0 & - & 38.5 & 46.2 & 40.0 & 43.6 & 42.1 & - & - \\ 
             & MMNet \cite{wu2021learning} & 34.9 & 41.0 & 37.2 & 37.0 & 37.5 & - & 37.0 & 40.3 & 39.3 & 36.0 & 38.2 & - & 10.4 M \\ 
             & MLC \cite{yang2021mining} & \textbf{46.8} & 35.3 & 26.2 & 27.1 & 33.9 & - & 54.1 & 41.2 & 34.1 & 33.1 & 40.6 & - & 8.7 M \\ 
             & CMN \cite{xie2021few} & 37.9 & \textbf{44.8} & 38.7 & 35.6 & 39.3 & 61.7 & 42.0 & 50.5 & 41.0 & 38.9 & 43.1 & 63.3 & - \\ 
             & HSNet \cite{hsnet} & 36.3 & 43.1 & 38.7 & 38.7 & 39.2 & 68.2 & 43.3 & \textbf{51.3} & \textbf{48.2} & 45.0 & 46.9 & 70.7 & 2.6 M \\   
             & \ccol \ournet & \ccol 41.5 & \ccol 44.1 & \ccol \textbf{42.8} & \ccol \textbf{40.6} & \ccol \textbf{42.2} & \ccol \textbf{68.8} & \ccol \textbf{47.6} & \ccol 50.1 & \ccol 47.7 & \ccol \textbf{46.4} & \ccol \textbf{47.9} & \ccol \textbf{71.6} & \ccol \textbf{1.3 M} \\
             \hline
             \multirow{7}{*}{R101} 
             & FWB \cite{nguyen2019fwb} & 17.0 & 18.0 & 21.0 & 28.9 & 21.2 & - & 19.1 & 21.5 & 23.9 & 30.1 & 23.7 & - & 43.0 M \\ 
             & DAN \cite{wang2020dan} & - & - & - & - & 24.4 & 62.3 & - & - & - & - & 29.6 & 63.9 & - \\ 
             & PFENet \cite{tian2020pfenet} & 34.3 & 33.0 & 32.3 & 30.1 & 32.4 & 58.6 & 38.5 & 38.6 & 38.2 & 34.3 & 37.4 & 61.9 & 10.8 M \\ 
             & SAGNN \cite{xie2021scale} & 36.1 & 41.0 & 38.2 & 33.5 & 37.2 & 60.9 & 40.9 & 48.3 & 42.6 & 38.9 & 42.7 & 63.4 & - \\ 
             & MLC \cite{yang2021mining} & \textbf{50.2} & 37.8 & 27.1 & 30.4 & 36.4 & - & \textbf{57.0} & 46.2 & 37.3 & 37.2 & 44.4 & - & 27.7 M\\ 
             & HSNet \cite{hsnet} & 37.2 & 44.1 & 42.4 & 41.3 & 41.2 & 69.1 & 45.9 & \textbf{53.0} & \textbf{51.8} & 47.1 & \textbf{49.5} & 72.4 & 2.6 M  \\  
             & \ccol \ournet & \ccol 41.8 & \ccol \textbf{45.4} & \ccol \textbf{43.2} & \ccol \textbf{41.9} & \ccol \textbf{43.1} & \ccol \textbf{69.4} & \ccol 48.0 & \ccol 52.1 & \ccol 49.7 & \ccol \textbf{48.2} & \ccol \textbf{49.5} & \ccol \textbf{72.7} & \ccol \textbf{1.3 M} \\          
             
             \bottomrule
        \end{tabular}
    }
    \vspace{-2mm}
    \caption{Fold-wise \fsstask results on 1-way 1-shot and 1-way 5-shot setups on COCO-20$^i$ using ResNet50 (R50) and ResNet101 (R101).}
    \label{table:fco50_foldwise}
    \vspace{-2mm}
\end{table*}
\smallbreakparagraph{Fold-wise results on COCO-$\mathbf{20^{i}}$.}
Tables~\ref{table:ico50_foldwise} and \ref{table:fco50_foldwise} present fold-wise performance comparison on the \ourtask and \fsstask tasks, respectively.
We validate that \ournet outperforms the competitors by large margins in both the \ourtask and \fsstask tasks on the challenging COCO-$20^{i}$ benchmark.

\smallbreakparagraph{Numerical performances of Fig.~4 in the main paper.}
We report the numerical performances of the Fig.~4 in the main paper in \tableref{table:multiway} as a reference for following research.

\end{document}